\documentclass[letterpaper]{article} 
\usepackage[preprint]{aaai2027}  
\usepackage[hyphens]{url}  
\usepackage{graphicx} 
\usepackage{natbib}  
\usepackage{caption} 
\title{Similar Models Learn Differently: Final Window Pretraining Shapes Post-Training Beyond SFT}
\author{
    Cen Lu\textsuperscript{\rm 1,2},
    Yung-Chen Tang\textsuperscript{\rm 1,2},
    Andrea Cavallaro\textsuperscript{\rm 1}
}
\affiliations{
    \textsuperscript{\rm 1}EPFL\\
    \textsuperscript{\rm 2}Idiap Research Institute
}
 
\copyrighttext{Preprint.}
 
\begin{document}

\maketitle

\begin{abstract}
Developers judge a model checkpoint by how it behaves. After supervised fine-tuning (SFT), two checkpoints that perform about the same across relevant benchmarks are treated as interchangeable, equally ready for the next alignment stage, typically preference optimization. We ask whether this judgment misses a pretraining imprint: a difference that no post-SFT benchmark reveals, yet that decides how each checkpoint responds to further training. To find out, we run a controlled experiment on the final window of pretraining, the last data a model is trained on before instruction tuning. Six branches fork from one partially pretrained checkpoint and differ only in this window: 500 million tokens, 0.1\% to 1\% of the pretraining tokens that precede it. Each branch trains its window on a single data source: generic web text, filtered web text, normative discourse, safety text, mathematical text, or synthetic educational text. SFT and post-training are then identical. After SFT the branches behave near-identically, within about one point on instruction following, refusal, and capability, yet the same post-training carries them to very different endpoints, under both a direct preference optimization update and a reinforcement learning update with a verifiable reward. We measure this deviation through refusal of harmful requests: when post-training begins the safety text branch refuses no more than the web text branch, yet by the end it has lost far less of its refusal. The other four branches gain little or no protection, so the effect is selective to what the window contained. The protection requires the safety text to arrive in the final window rather than earlier in pretraining, and it reproduces on a second model family. What a model is pretrained on last shapes how it reacts to alignment. Therefore, a checkpoint should not be evaluated by its post-SFT behavior alone, and what it was trained on last should be reported with it.
\end{abstract}
\section{Introduction}

Large language models are built in stages, starting with pretraining for base knowledge, followed by supervised fine-tuning (SFT) and preference optimization such as Direct Preference Optimization (DPO) to turn the base model into a helpful assistant~\citep{ouyang2022training,rafailov2023direct}. Developers decide whether an SFT checkpoint is ready for the next stage purely by testing its behavior: whether it follows instructions, safely refuses harmful prompts, and maintains its overall capabilities~\citep{tulu2023}. When two checkpoints pass these tests equally well, nothing in the evaluation tells them apart.

\begin{figure*}[t]
\centering
\includegraphics[width=\textwidth]{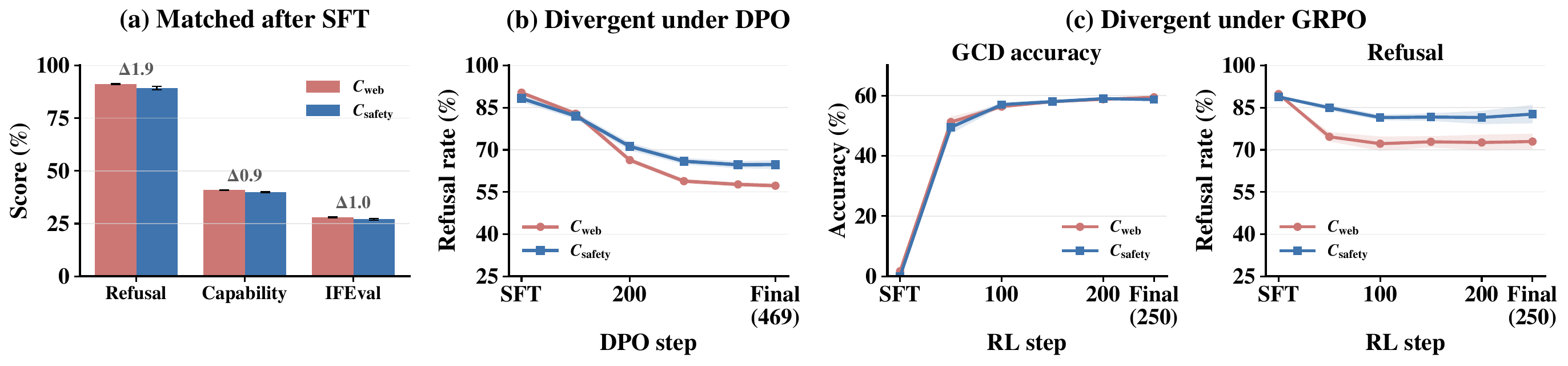}
\caption{Similar models learn differently. Two branches differ only in their final pretraining window, $C_{\mathrm{web}}$ on generic web text and $C_{\mathrm{safety}}$ on safety data, forking from the same 49B token OLMo-2-1B checkpoint with a 500M token final window, then running identical Tulu-style SFT and post-training over three seeds. \textbf{(a)} After SFT the branches are closely matched on refusal, capability, and instruction following (Instruction Following Evaluation, IFEval). Each pair differs by about one point and $C_{\mathrm{safety}}$ is never ahead. \textbf{(b)} Under the same UltraFeedback DPO their overall refusal, averaged over AdvBench, BeaverTails, and XSTest-unsafe, diverges, with $C_{\mathrm{safety}}$ starting no higher yet losing less and ending higher. \textbf{(c)} The same divergence also holds under a GRPO reinforcement learning update on GCD, the greatest common divisor of two integers, where both branches learn the task equally while $C_{\mathrm{safety}}$ loses far less refusal. Bands are three-seed standard deviations. }
\label{fig:erosion-curve}
\end{figure*}

However, judging a model only by its final behavior ignores the path that training took to get there. That path leaves a lasting mark on how a model adapts: models that share a trajectory converge to the same region of the loss landscape~\citep{frankle2020linear,neyshabur2020being}, and plasticity, the capacity of a network to keep changing under further training, is itself shaped by how it was trained~\citep{lyle2023understanding,dohare2024loss}. In language models, small late stage pretraining mixtures change downstream fine-tuning behavior~\citep{baek2026finetuners,feng2026early}, and front-loading reasoning data into pretraining changes what post-training can reach~\citep{akter2026frontloading}. Past work therefore establishes that the training path matters, but none of it tests whether two checkpoints matched in behavior after SFT still carry an imprint of that path which decides how they respond to further training.

Our hypothesis is that the final window of pretraining, the last data trained on before instruction tuning, is what sets the imprint. We track it through refusal of harmful requests, because refusal is installed by a component of SFT that we hold fixed across branches and is easy to measure across many prompts. Starting from one partially pretrained OLMo-2-1B checkpoint~\citep{olmo22025}, we fork six branches that differ only in that window, and then apply identical Tulu-style SFT~\citep{tulu2023} and UltraFeedback DPO~\citep{ultrafeedback2023} to every branch, adding a Group Relative Policy Optimization (GRPO) reinforcement learning (RL) update on the two headline branches. We measure refusal erosion, the drop in refusal from the SFT checkpoint to the post-training checkpoint, which isolates the change caused by the shared post-training stage. Erosion separates the two quantities that a comparison of endpoints confuses: how much refusal a branch has when post-training begins, and how much of it the same update removes. The question is therefore not whether some pretraining data makes a model safer, but whether the pretraining path, both what the final window contained and the fact that it came last, decides how much of what SFT installed survives the next stage. Figure~\ref{fig:erosion-curve} previews the answer: closely matched on the measured post-SFT behavior, the branches diverge under the same DPO, and the same divergence appears under GRPO.

We first verify that the branches are matched after SFT and diverge under the same DPO. We then ask what drives the divergence: which final window content protects refusal, and whether that content must come last. We then generalize beyond DPO to the GRPO update, and test whether the effect is robust across the learning rate, the preference data, and a second model family (Pythia-1B~\citep{biderman2023pythia}), and how it weakens as the final window becomes a smaller fraction of prior training. That last test varies the fraction two ways: fixing the 500M window and forking later, and, as a control, shrinking the window at a fixed fork point.

We summarize our contributions as follows. First, we identify a path dependence of post-training: checkpoints matched on the standard post-SFT criteria of instruction following, refusal, and capability still diverge under an identical post-training update, so these criteria are blind to a difference set by late pretraining. Second, we isolate the effect with a refusal erosion metric and a matched checkpoint control, and we show that the safety branch loses less refusal under post-training than the web branch despite not starting higher. Third, under DPO we characterize when and how strongly the effect holds: it depends on the final window content, requires the safety data to appear last, weakens to near zero as the window shrinks to a few basis points of prior training, and replicates on a different model family. The divergence is not specific to one algorithm: it also appears under a GRPO reinforcement learning (RL) update with verifiable rewards. A model's plasticity for alignment is not fully visible in its post-SFT behavior, so downstream developers that align a model need to know what it was trained on last, not just how it scores.

\section{Related Work}

\paragraph{Pretraining data and training stage.}
Scaling law and computationally optimal training work established that model behavior depends on both model size and data scale~\citep{kaplan2020scaling,hoffmann2022training}. Open model suites and large pretraining corpora made it possible to study this dependence with checkpoints, curated datasets, and controlled recipes~\citep{biderman2023pythia,groeneveld2024olmo}. Recent data selection work further shows that the composition and quality of pretraining data can change downstream behavior, not only perplexity~\citep{li2024datacomp}. More directly, midtraining uses synthetic documents about desired behavior after pretraining and before alignment fine-tuning, showing that an intermediate stage can affect how later alignment data generalizes~\citep{li2026modelspec}. These lines of work motivate our experimental design, but instead of using better pretraining data to improve a model in general, we study whether post-training is path-dependent on the final window of continued pretraining, eroding an SFT-installed behavior by different amounts across otherwise matched branches. Additionally, small data fractions can also matter during pretraining. Specialized pretraining mixes target domain data into general pretraining at small mixture rates, including percent scale settings such as 1--2\%, and changes later fine-tuning outcomes~\citep{baek2026finetuners}. Early data exposure experiments similarly show that mixing target data into pretraining can affect how robustly a later post-trained capability survives subsequent fine-tuning~\citep{feng2026early}. These results support treating a small final window pretraining intervention as measurable rather than negligible.

\paragraph{Alignment oriented pretraining text.}
Some alignment oriented work is already textual before it becomes a post-training objective: constitutions, preference feedback, and user value surveys all describe intended assistant behavior or human judgments in natural language~\citep{bai2022constitutional,kirk2024prism,openassistant2023,wang2024helpsteer2}. This makes it plausible that late pretraining on such material could change the state entering post-training. Our experiment uses them to separate normative discourse from safety transformation text, then tests whether either changes a later post-training erosion dynamic under a fixed recipe.

\paragraph{Post-training, preference optimization, and refusal.}
Instruction tuning and reinforcement learning from human feedback (RLHF) made post-training a standard mechanism for converting pretrained models into assistants~\citep{ouyang2022training,bai2022constitutional}. DPO provides a simpler preference optimization objective that is now widely used in open post-training pipelines~\citep{rafailov2023direct,tulu2023,ultrafeedback2023}. Safety evaluations such as AdvBench, XSTest, BeaverTails, and OR-Bench measure harmful request refusal and over-refusal from different angles~\citep{zou2023advbench,rottger2023xstest,ji2023beavertails,orbench2024}. We use these tools as probes to study whether the same post-training recipe, DPO or an RL update, erodes a refusal prior installed by SFT differently after different final pretraining windows. A related line studies how alignment itself can be made robust to subsequent fine-tuning that degrades refusal, for example perturbation aware alignment that produces hidden embeddings invariant to harmful fine-tuning~\citep{huang2024vaccine}. That work intervenes at the alignment stage against adversarial fine-tuning, whereas we ask whether a pretraining choice, the final window corpus, changes how much a refusal prior erodes under standard benign post-training.

\begin{table*}[t]
\centering
\small
\setlength{\tabcolsep}{4pt}
\begin{tabular}{@{}p{0.08\textwidth}p{0.58\textwidth}p{0.12\textwidth}p{0.12\textwidth}@{}}
\hline
Branch & Final window corpus & Unique tokens & Main use \\
\hline
$C_{\mathrm{web}}$ & FineWeb-Edu Web text~\citep{penedo2024fineweb} & 16.7B & sample 500M \\
$C_{\mathrm{dclm}}$ & DCLM-baseline Web text~\citep{li2024datacomp} & 635M & sample 500M \\
$C_{\mathrm{norm}}$ & Normative Discourse transform~\citep{openai2025modelspec,kirchner2022understanding,kim2024prometheus,kirk2024prism,openassistant2023,wang2024helpsteer2,bai2022helpful} & 396M & 1.26 passes \\
$C_{\mathrm{safety}}$ & Safety Transformation text~\citep{ji2023beavertails,ji2025pku,bai2022helpful,lin2023toxicchat,gehman2020realtoxicityprompts,hendrycks2021ethics} & 219M & 2.3 passes \\
$C_{\mathrm{math}}$ & OpenWebMath Math Web text~\citep{paster2023openwebmath} & 440M & 1.14 passes \\
$C_{\mathrm{synth}}$ & Cosmopedia Synthetic Education text~\citep{benallal2024cosmopedia} & 1.05B & sample 500M \\
\hline
\end{tabular}
\caption{Final window corpora used in the main experiment. The token budget is fixed at 500M tokens, smaller corpora are cycled to match the budget. $C_{\mathrm{math}}$ is mathematical web text, while $C_{\mathrm{synth}}$ is synthetic education. $C_{\mathrm{math}}$ and $C_{\mathrm{synth}}$ are two specialized non-safety controls, not a shared data family.}
\label{tab:corpora-main}
\end{table*}

\section{Methodology}

Our goal is to measure whether the same post-training update erodes an SFT-installed behavior differently in two branches it moves equally overall, that is, whether post-training is path-dependent on the final pretraining window separating them. Let $\theta_0$ be a shared pretrained checkpoint. For each final window corpus $D_c$, we apply continued pretraining (CPT) for a fixed token budget $T$:
\begin{equation}
\theta_c = \mathrm{CPT}(\theta_0, D_c, T).
\end{equation}
We then apply the same post-training to every branch, supervised fine-tuning followed by a post-training update $\mathrm{PT}$:
\begin{equation}
\theta_c^S = \mathrm{SFT}(\theta_c, D_{\mathrm{sft}}), \quad
\theta_c^{\mathrm{PT}} = \mathrm{PT}(\theta_c^S).
\end{equation}
Here $\theta_c^{\mathrm{PT}}$ is the post-training endpoint under whichever update is applied. We use two updates: UltraFeedback DPO, our primary instance, and a GRPO reinforcement learning update that maximizes a verifiable reward under a KL penalty to the SFT policy (Section~\ref{sec:not-dpo}). DPO uses the standard pairwise preference objective~\citep{rafailov2023direct}:
\begin{equation}
\begin{array}{l}
\mathcal{L}_{\mathrm{DPO}}(\theta) =
-\mathrm{E}_{(x,y^+,y^-)}
\log \sigma \Bigl(\beta[
\log \frac{\pi_\theta(y^+|x)}{\pi_{\mathrm{ref}}(y^+|x)}
\\
\hfill
- \log \frac{\pi_\theta(y^-|x)}{\pi_{\mathrm{ref}}(y^-|x)}
]\Bigr).
\end{array}
\end{equation}
Here $x$ is a prompt, $y^+$ and $y^-$ are the preferred and dispreferred responses, $\pi_\theta$ is the policy being trained, $\pi_{\mathrm{ref}}$ is the frozen reference policy (the SFT checkpoint $\theta_c^S$), $\sigma$ is the logistic function, and $\beta$ controls how far the policy can move from the reference.

\paragraph{Refusal erosion.}
Endpoint refusal is not enough for our question. A branch can end with higher refusal because it started higher after SFT, or because it lost less refusal under post-training. We define refusal erosion explicitly. For benchmark $b$ with prompts $B_b$,
\begin{equation}
R_b(\theta)=\frac{1}{|B_b|}\sum_{x\in B_b}\mathbf{1}\{\theta \textrm{ refuses } x\}.
\end{equation}
The erosion caused by a post-training update is
\begin{equation}
E_b(c)=R_b(\theta_c^S)-R_b(\theta_c^{\mathrm{PT}}).
\end{equation}
We compare branches through protection relative to the web baseline $C_{\mathrm{web}}$:
\begin{equation}
P_b(c)=E_b(C_{\mathrm{web}})-E_b(c).
\end{equation}
Positive $P_b(c)$ means branch $c$ loses less refusal than $C_{\mathrm{web}}$ under the same post-training stage.

This definition makes the comparison stage-aware. A high post-training refusal endpoint can come from a high SFT starting point, slow erosion, or both. We report SFT endpoints, post-training endpoints, erosion, and protection where space allows, but use erosion as the main quantity because it isolates the change caused by the shared post-training stage. We further verify that the branches are closely matched after SFT on refusal, instruction following, and capability (Figure~\ref{fig:erosion-curve}(a)), so a difference in erosion reflects how the shared stage moves each branch rather than a difference in starting point.

\paragraph{Branches and post-training.}
We fork from an early OLMo-2-1B checkpoint at roughly 49B training tokens. The main final window branches use matched 500M token continued pretraining windows: $C_{\mathrm{web}}$ Web, $C_{\mathrm{dclm}}$ DCLM, $C_{\mathrm{norm}}$ Normative, $C_{\mathrm{safety}}$ Safety, $C_{\mathrm{math}}$ Math, and $C_{\mathrm{synth}}$ Synth Edu. The safety transformation branch is not a corpus of bare refusal templates: it contains harmful scenarios, unsafe draft responses, and critique or revision text. After continued pretraining, every branch receives the same Tulu-style SFT and UltraFeedback DPO pipeline. The main 49B fork branches use three SFT/DPO/evaluation seeds.

The fork and window size are a stage-aware probe, not an estimate of an optimal pretraining recipe. The 49B token fork is early relative to trillion token open model training but past the smallest computationally optimal scale for a 1B model~\citep{hoffmann2022training}. The 500M token intervention is about 1\% of the training history at the fork, in the same percent scale range where small target data mixtures are known to affect later pretraining and fine-tuning behavior~\citep{baek2026finetuners,feng2026early}, while staying small enough to test final window effects rather than replacing the whole pretraining history.

Table~\ref{tab:corpora-main} gives the corpus level design, chosen to separate content from generic extra training. $C_{\mathrm{web}}$ and $C_{\mathrm{dclm}}$ are web text controls with different filtering, so $C_{\mathrm{dclm}}$ asks whether another high quality web filter changes the same post-training behavior. $C_{\mathrm{norm}}$ and $C_{\mathrm{safety}}$ are both about alignment but differ in content: $C_{\mathrm{norm}}$ is normative and model governance prose drawn from model spec, alignment discourse, preference feedback, and value survey sources~\citep{openai2025modelspec,kirchner2022understanding,kim2024prometheus,kirk2024prism,openassistant2023,wang2024helpsteer2,bai2022helpful}, while $C_{\mathrm{safety}}$ is harmful request scenarios with unsafe drafts and safety critique drawn from safety preference, toxicity, red teaming, and ethics sources~\citep{ji2023beavertails,ji2025pku,bai2022helpful,lin2023toxicchat,gehman2020realtoxicityprompts,hendrycks2021ethics}. $C_{\mathrm{math}}$ (OpenWebMath,~\citep{paster2023openwebmath}) and $C_{\mathrm{synth}}$ (Cosmopedia synthetic education,~\citep{benallal2024cosmopedia}) are non-safety controls with specialized content. Together with the fixed token budget, this lets us ask which final window changes the response to a shared post-training recipe, not which corpus is best for pretraining in general.

The post-training data is held fixed across branches. SFT uses a 100k example Tulu-style instruction subset from the training split~\citep{tulu2023}. This mixture includes a safety and refusal component that is held identical across branches, so the refusal prior we study is installed by SFT rather than by the final window corpus. DPO uses 60k UltraFeedback training preference pairs~\citep{ultrafeedback2023}. Held out UltraFeedback pairs are used only for the DPO proxy diagnostic (Appendix Table~\ref{tab:appendix-proxy}), not for training. We use the same DPO reference construction, optimizer settings, and checkpoint schedule for every branch, so the measured differences are downstream of the final window corpus rather than the post-training data. The GRPO reward and its train/evaluation split are likewise fixed across the two headline branches.

\paragraph{Order and stage controls.}
The word ``final'' is part of the hypothesis. To isolate order, we train two branches with the same $C_{\mathrm{dclm}}$ and $C_{\mathrm{safety}}$ content in opposite orders:
\begin{equation}
\theta_{\mathrm{dclm}\rightarrow\mathrm{safety}}=\mathrm{CPT}(\mathrm{CPT}(\theta_0,D_{\mathrm{dclm}},T),D_{\mathrm{safety}},T),
\end{equation}
\begin{equation}
\theta_{\mathrm{safety}\rightarrow\mathrm{dclm}}=\mathrm{CPT}(\mathrm{CPT}(\theta_0,D_{\mathrm{safety}},T),D_{\mathrm{dclm}},T).
\end{equation}
Both see the same content and total token budget, only position differs. We also run a stage boundary check from a saturated OLMo-2-1B checkpoint at roughly 4T training tokens. The same absolute 500M token intervention is about 1\% of the early fork's prior training, but only a very small relative intervention at the saturated fork.

\paragraph{Evaluations.}
We measure harmful request refusal on AdvBench~\citep{zou2023advbench}, BeaverTails~\citep{ji2023beavertails}, and XSTest-unsafe~\citep{rottger2023xstest}, over-refusal on OR-Bench~\citep{orbench2024}, instruction following on IFEval~\citep{zhou2023instructionfollowing}, and capability on MMLU~\citep{hendrycks2021mmlu}, HellaSwag~\citep{zellers2019hellaswag}, ARC-Challenge~\citep{clark2018arc}, and WinoGrande~\citep{sakaguchi2020winogrande}. Since BeaverTails has prompt overlap with the safety transformation source data, we report it as contaminated and use decontaminated BeaverTails as supporting evidence. We report accuracy (acc) and normalized accuracy (acc-norm) for capability checks, where acc-norm is the length normalized multiple choice score from the evaluation harness. Persona, value, and sycophancy probes (Appendix~\ref{app:fragility}), decontamination details (Appendix~\ref{app:decontam}), and corpus diagnostics (Appendix~\ref{app:corpus}) are shown in the appendix. All branches use the same generation and refusal classification path on the same prompts, so the controlled branch to branch comparison is the relevant quantity.

\paragraph{Refusal measurement.}
We classify a completion as a refusal using lexical refusal string matching,
following the AdvBench protocol~\citep{zou2023advbench}. A response counts as a
refusal if any of a fixed list of ten refusal patterns (e.g., \emph{I cannot},
\emph{I'm sorry}) matches case insensitively within the first 300 characters of the
generated text, i.e., the refusal prefix region. The same patterns and generation
path are applied to every branch and benchmark, so absolute refusal rates depend on
this detector but branch to branch differences remain comparable. Because lexical
matching can be fooled by refusal style prefixes that still comply, we validate it
against the non-lexical WildGuard classifier~\citep{han2024wildguard} in
Appendix~\ref{app:wildguard}: the two agree on $97$--$99\%$ of completions, the
$C_{\mathrm{web}}$--$C_{\mathrm{safety}}$ protection reproduces under WildGuard, with a mean of $+8.2$ points
across the five validation benchmarks versus keyword $+9.3$ on the same completions, and $C_{\mathrm{safety}}$'s keyword false positive rate is within $0.7$ points of $C_{\mathrm{web}}$'s.

\section{Experiments}

In this section, we first show the branches are matched after SFT yet diverge under the same DPO. We then test which properties of the final window drive it, its content and then its position, generalize beyond DPO to a GRPO RL update, and check whether the effect is robust across recipe and model and where it breaks down.

\subsection{Matched After SFT, Divergent Under DPO}

After SFT, the safety branch does not start higher than the web baseline on harmful refusal benchmarks, and the two branches are closely matched overall on refusal, capability, and instruction following (Figure~\ref{fig:erosion-curve}(a)). Under the same UltraFeedback DPO stage, however, the web branch loses refusal faster. Figure~\ref{fig:erosion-curve} shows this with dense checkpoints: in overall refusal, $C_{\mathrm{safety}}$ starts below $C_{\mathrm{web}}$ after SFT, crosses it during DPO, and ends higher because its refusal erodes slowly.

The lower SFT starting point matters for interpretation. If $C_{\mathrm{safety}}$ started with higher refusal after SFT, a higher post-DPO endpoint could be a ceiling effect or a stronger initial refusal policy. Instead, the safety branch starts at a disadvantage on the dense overall trajectory and still loses less refusal under the same DPO update. This result shows a difference in post-training susceptibility, not evidence that safety final window data induces stronger refusal after SFT.

Table~\ref{tab:main-erosion} summarizes the three-seed erosion result on different benchmarks. Across all benchmarks, $C_{\mathrm{safety}}$ loses less refusal than $C_{\mathrm{web}}$. The strongest evidence comes from AdvBench and XSTest, which barely overlap the $C_{\mathrm{safety}}$ source. All numbers are decontaminated (overlapping evaluation prompts removed), so the BeaverTails column reflects clean results. The gap differs across benchmarks as expected: AdvBench, a cleaner out-of-domain adversarial set, shows a moderate gap, XSTest-unsafe the largest, and BeaverTails, which overlaps the $C_{\mathrm{safety}}$ source, is reported decontaminated throughout (overlap audit in Appendix~\ref{app:decontam}).

\begin{table}[t]
\centering
\footnotesize
\setlength{\tabcolsep}{4pt}
\begin{tabular}{lccc}
\hline
Benchmark & $E_b(C_{\mathrm{web}})$ & $E_b(C_{\mathrm{safety}})$ & $P_b(C_{\mathrm{safety}})$ \\
\hline
AdvBench & $11.7\pm0.3$ & $8.4\pm0.2$ & $3.3\pm0.5$ \\
BeaverTails & $33.0\pm1.5$ & $27.8\pm0.6$ & $5.2\pm1.2$ \\
XSTest-unsafe & $51.9\pm1.2$ & $35.7\pm0.6$ & $16.2\pm1.2$ \\
\hline
Overall & $32.2\pm0.8$ & $24.0\pm0.1$ & $8.2\pm0.9$ \\
\hline
\end{tabular}
\caption{DPO refusal erosion for the $C_{\mathrm{web}}$ and $C_{\mathrm{safety}}$ branches, on prompts decontaminated against the $C_{\mathrm{safety}}$ source data (overlapping evaluation prompts removed, Appendix~\ref{app:decontam}). For benchmark $b$, $E_b(c)=R_b(\theta_c^S)-R_b(\theta_c^D)$ is the refusal lost from SFT to DPO (SFT minus DPO refusal rate), and $P_b(C_{\mathrm{safety}})=E_b(C_{\mathrm{web}})-E_b(C_{\mathrm{safety}})$ is $C_{\mathrm{safety}}$'s protection relative to $C_{\mathrm{web}}$. Values are three-seed means in percentage points with seed standard deviations, the protection standard deviation is paired across seeds. Lower erosion is better, positive protection means $C_{\mathrm{safety}}$ loses less refusal than $C_{\mathrm{web}}$ under the same DPO stage.}
\label{tab:main-erosion}
\end{table}

\subsection{The Divergence Is Selective to Final Window Content}

We now ask what drives the divergence, beginning with the content of the final window. All six final window branches leave the SFT checkpoint statistically indistinguishable from $C_{\mathrm{safety}}$; under the same DPO they separate significantly except $C_{\mathrm{synth}}$ (Welch's $t$-tests, Appendix Table~\ref{tab:allbranch-ttest}; per-branch trajectories in Appendix Figure~\ref{fig:all-dpo-curves}). Which branches gain protection and how much is what this subsection characterizes.

\begin{figure*}[tp]
\centering
\includegraphics[width=.92\textwidth]{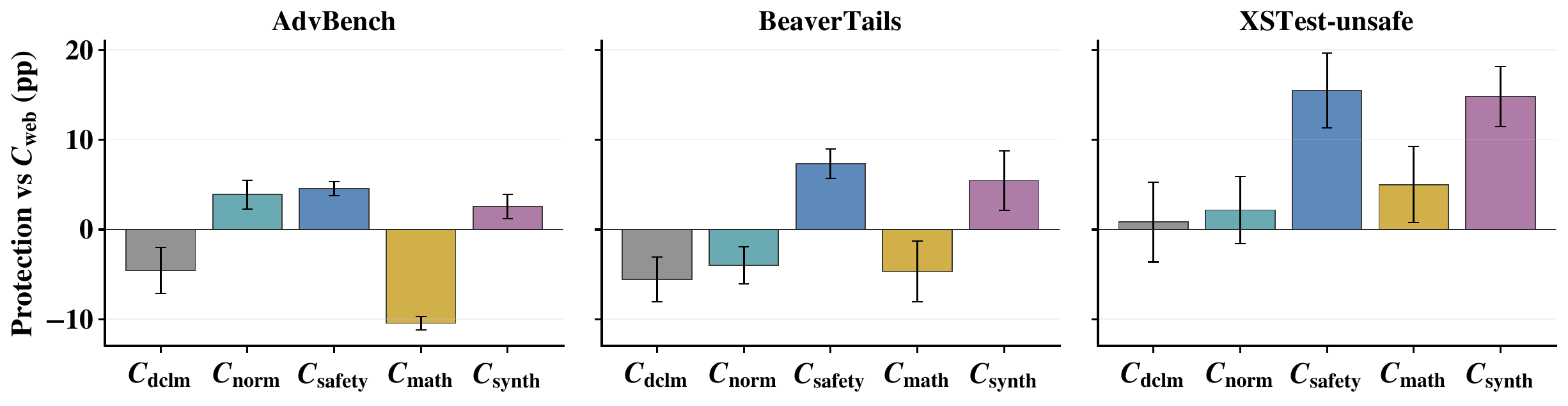}
\caption{Protection against DPO erosion relative to $C_{\mathrm{web}}$. Higher protection means less refusal is lost relative to $C_{\mathrm{web}}$ under the same update. $C_{\mathrm{safety}}$ is the most consistent branch across harmful refusal benchmarks, other corpora show partial or benchmark specific effects. All branches start from a matched SFT refusal level, so this is resistance to erosion rather than a higher starting refusal.}
\label{fig:protection}
\end{figure*}

The result is not that any late continued pretraining stabilizes refusal. Figure~\ref{fig:protection} compares protection relative to the web baseline across all final window corpora. DCLM and math are not protective. Normative discourse is close to null overall: it is mildly positive on AdvBench and XSTest but negative on BeaverTails, so it does not give a consistent cross benchmark effect. Synthetic education is the one non-safety branch with a clear partial effect: strong on XSTest, positive on BeaverTails, but weak on AdvBench. Neither the normative nor the synthetic branch matches the cross benchmark pattern of safety transformation text. This selectivity matters for the scope of the paper: the result is a content dependent change in post-training susceptibility, not a generic benefit from extra tokens.

The selective pattern argues against a simple refusal template explanation. The normative branch contains more explicit refusal language than the safety branch, but it does not reproduce the same robust erosion protection. The safety branch is better described as harmful scenario and safety critique heavy. This points away from mimicry and toward a task proximal prior that interacts with later post-training.

The DCLM branch is also informative. $C_{\mathrm{dclm}}$ is not a degraded continuation corpus, it is another web derived corpus with a different filter. If the main effect were caused by exposing the model to higher quality text near the fork point, $C_{\mathrm{dclm}}$ should move in the same direction as $C_{\mathrm{safety}}$, which it does not. This makes $C_{\mathrm{web}}$ and $C_{\mathrm{dclm}}$ a pair of web like controls. The math branch provides a different negative control: non-safety text can change capabilities and language statistics, but it does not protect refusal in the same way.

\subsection{The Protective Content Must Come Last}

\begin{figure}[!htbp]
\centering
\includegraphics[width=\linewidth]{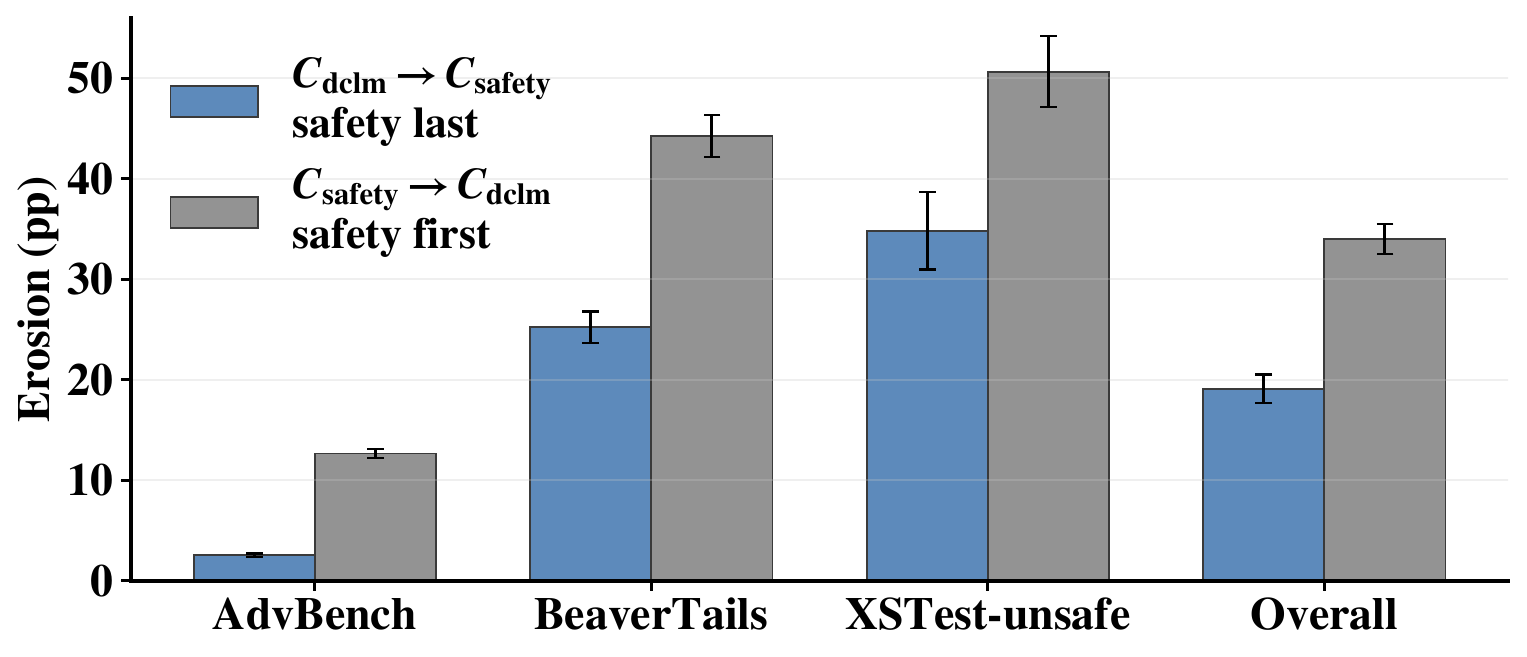}
\caption{Order control erosion over three seeds. Both branches see the same DCLM and safety content. Lower erosion is better. Placing safety data last yields lower erosion, placing it first weakens the effect, so what the model sees last matters, not merely that it saw safety data during pretraining.}
\label{fig:order}
\end{figure}

Content is necessary but not sufficient: its position matters too. The order control tests whether safety exposure somewhere in late pretraining is enough. We pair $C_{\mathrm{safety}}$ with $C_{\mathrm{dclm}}$, a high quality web like continuation that is not protective in the main grid, so we can ask whether safety content must be last after a non-safety window without tying the order effect to the exact web corpus. Figure~\ref{fig:order} reports the three-seed result. With identical DCLM and safety content, the safety last branch has lower DPO erosion across all three harmful refusal benchmarks, while the safety first branch washes out toward the web and DCLM controls. The mean erosion is $2.6/25.2/34.8$ for safety last versus $12.7/44.2/50.7$ for safety first on AdvBench, BeaverTails, and XSTest-unsafe. This makes final window position an important ingredient in our setting.

Within this continued pretraining intervention, the final position of safety transformation text matters for how DPO later erodes refusal. That distinction rules out a simpler exposure only story while avoiding claims about arbitrary training scales or data orders. $C_{\mathrm{dclm}}$ by itself is not protective in the main grid, and $C_{\mathrm{safety}}$ is protective. If the order control were only measuring total exposure to safety examples, both $C_{\mathrm{dclm}}\rightarrow C_{\mathrm{safety}}$ and $C_{\mathrm{safety}}\rightarrow C_{\mathrm{dclm}}$ would be similar. Empirically, the later window changes the subsequent DPO response more strongly than the earlier window in this setting. This is why we claim the intervention as final window pretraining rather than pretraining data exposure. More broadly, this control also reduces a size and content confound: it holds the two corpora and total token budget fixed, so only placement changes, yet the safety last branch preserves refusal while the safety first branch tracks the web like controls.

\subsection{The Same Divergence Appears Under RL}
\label{sec:not-dpo}

The divergence also generalizes beyond DPO. We run RL with GRPO from the same $C_{\mathrm{web}}$ and $C_{\mathrm{safety}}$ SFT checkpoints, using a verifiable reward on the greatest common divisor (GCD) of two integers from $1$ to $1000$ ($4000$ training and $400$ held out evaluation instances). Both branches learn the task equally, rising from near zero to $59.4\pm0.2$ and $58.7\pm0.5$ percent exact accuracy over three seeds along overlapping trajectories (Figure~\ref{fig:erosion-curve}(c), left), so a refusal difference cannot come from how much the update moves each branch. Yet refusal diverges under the same update (Figure~\ref{fig:erosion-curve}(c), right): $C_{\mathrm{web}}$ loses $16.9\pm2.8$ points of overall refusal while $C_{\mathrm{safety}}$ loses $6.1\pm3.3$, a protection of $10.8\pm1.0$ points. This is the same matched trainability with divergent refusal we see under DPO, and a second RL task, Countdown arithmetic, reproduces it (Appendix~\ref{app:rl-countdown}), so the effect is a property of post-training in general rather than of one algorithm. Neither update targets safety: DPO optimizes preference for helpfulness and GRPO a correct arithmetic answer, so in both the refusal loss is a side effect, and the divergence reflects how differently two matched branches give up a behavior the update never rewards.

\subsection{The Divergence Survives Recipe and Model}

\begin{figure*}[tp]
\centering
\includegraphics[width=.78\textwidth]{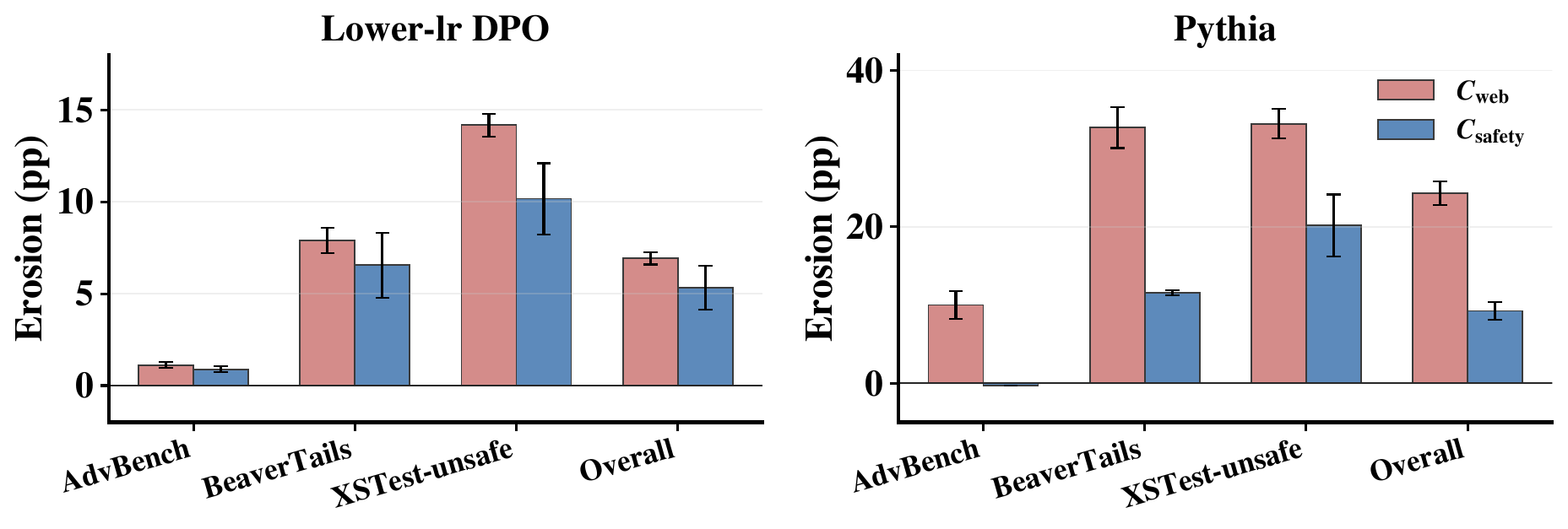}
\caption{Robustness checks for the $C_{\mathrm{web}}$--$C_{\mathrm{safety}}$ erosion contrast (bars show DPO erosion, lower is better). Left: lower learning rate DPO, $2{\times}10^{-7}$ versus the main $5{\times}10^{-7}$, on the OLMo 49B fork branches. Right: the same $C_{\mathrm{web}}$/$C_{\mathrm{safety}}$ final window contrast on Pythia-1B-deduped~\citep{biderman2023pythia}, a different model forked at checkpoint step 24{,}000 (about 50B tokens, close to the OLMo 49B fork). Overall is the mean over AdvBench, BeaverTails, and XSTest-unsafe.}
\label{fig:robustness-main}
\end{figure*}

The contrast is robust to how the final window is built, whether one is added at all, and whether we decontaminate the eval: it survives collapsing the repeated $C_{\mathrm{safety}}$ corpus to a single pass (Appendix Figure~\ref{fig:repetition}), persists for a no-continuation Base branch with the same SFT and DPO (Appendix~\ref{app:base-control}), and holds on decontaminated prompts, with AdvBench and XSTest at near-zero overlap and the BeaverTails audit in Appendix~\ref{app:decontam}.

The contrast also holds across settings (Figure~\ref{fig:robustness-main}). Lowering the DPO learning rate from $5{\times}10^{-7}$ to $2{\times}10^{-7}$ reduces erosion for both branches but preserves the gap. Repeating the $C_{\mathrm{web}}$/$C_{\mathrm{safety}}$ contrast on Pythia-1B-deduped at the same 50B token region~\citep{biderman2023pythia}, a different model, reproduces the direction: mean overall erosion is $25.3$ points for $C_{\mathrm{web}}$ versus $10.5$ for $C_{\mathrm{safety}}$, with lower $C_{\mathrm{safety}}$ erosion on all three benchmarks (Appendix Table~\ref{tab:pythia-external}). The SFT starting points differ across branches, so we treat Pythia as supporting evidence rather than a replacement for the controlled OLMo grid. The protection also survives changing the preference dataset from UltraFeedback to human Chatbot Arena votes (Appendix~\ref{app:pref-robustness}).

\subsection{Boundaries and Costs}

\begin{figure}[t]
\centering
\includegraphics[width=.95\columnwidth]{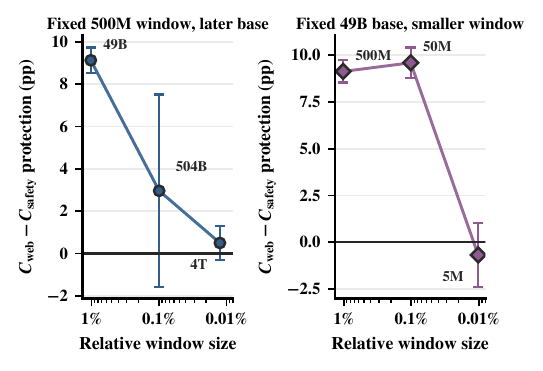}
\caption{Relative dose attenuation for the overall $C_{\mathrm{web}}$--$C_{\mathrm{safety}}$ erosion contrast. Protection is $C_{\mathrm{web}}$ erosion minus $C_{\mathrm{safety}}$ erosion, so higher is better. Left: fixed 500M token final window at progressively later OLMo checkpoints, the actual relative doses are $1.02\%$, $0.099\%$, and $0.0125\%$. Right: fixed 49B token checkpoint with 500M, 50M, and 5M final windows, the actual relative doses are $1.02\%$, $0.102\%$, and $0.010\%$. The fixed checkpoint control shows that a final window at around $0.1\%$ -- $1\%$ relative dose is enough to drive two otherwise matched branches apart under DPO, and the gap fades at smaller doses.}
\label{fig:relative-dose-main}
\end{figure}

The protection is bounded by relative dose: it weakens as the same 500M token window becomes a smaller fraction of prior training and fades to near zero at basis point scale (Figure~\ref{fig:relative-dose-main}). The dose dependence is itself informative. The fixed 500M token window spans roughly three relative dose orders of magnitude: $1.02\%$ at the 49B token fork, $0.099\%$ at 504B, and $0.0125\%$ at the saturated 4.001T token endpoint. Overall protection, a benchmark mean on raw prompts, falls from $9.1$ points at 49B to $3.0$ at 504B (paired seed standard deviation $4.5$ points) and about $0.5$ at the saturated 4.001T checkpoint, whose post-DPO endpoints are in Appendix Table~\ref{tab:boundary}. A fixed 49B control agrees, a 5M window ($\approx0.010\%$) is near null while a 50M window ($\approx0.102\%$) recovers a clear signal. The relative dose thus sets how far apart two SFT-matched branches end up under the same post-training.

\begin{figure}[t]
\centering
\includegraphics[width=.8\columnwidth]{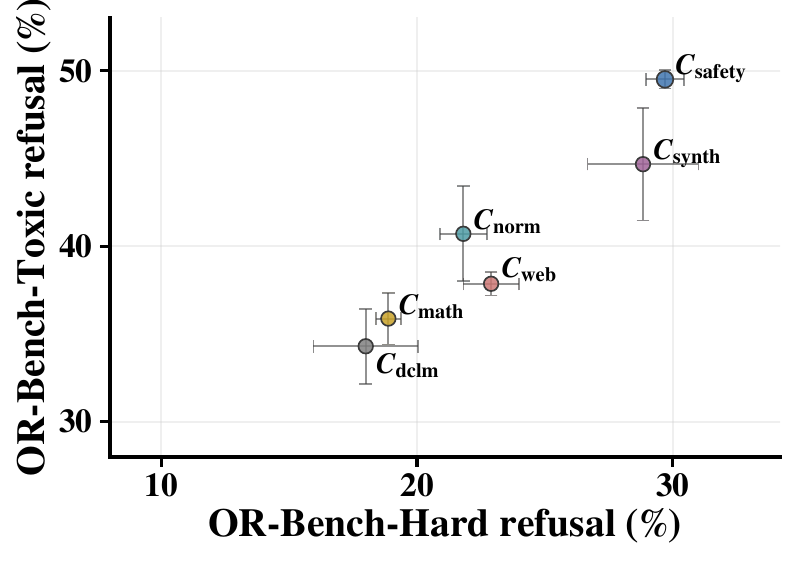}
\caption{OR-Bench refusal tradeoff~\citep{orbench2024}. Points are three-seed means, error bars are seed standard deviations. The horizontal axis is refusal on OR-Bench-Hard safe or borderline prompts, where lower is better, the vertical axis is refusal on OR-Bench-Toxic prompts, where higher is better. $C_{\mathrm{safety}}$ refuses more on both, so the retained refusal is a broad prior rather than calibrated harm refusal.}
\label{fig:overrefusal-main}
\end{figure}

The divergence is also a broad shift in the refusal prior rather than a calibrated gain in harmful request refusal. $C_{\mathrm{safety}}$ over-refuses benign prompts as well, raising OR-Bench-Hard over-refusal above $C_{\mathrm{web}}$ (Figure~\ref{fig:overrefusal-main}), and it carries a small capability cost of about $0.9$ points on a four task mean, already present at the SFT checkpoint rather than opened by DPO (Appendix Tables~\ref{tab:appendix-proxy} and~\ref{tab:appendix-utility}). What it retains is refusal behavior rather than lower measured output harm, since a harm classifier (WildGuard~\citep{han2024wildguard}) shows at most a weak difference between the two on the harm axis.

\section{Conclusion}

A small final window of pretraining can change how the next stage of alignment reshapes a model. In our setting, safety text in that window does not make the model refuse more after SFT; what it changes is how much of that refusal survives the next update. The safety branch loses far less under the same DPO, and the same divergence appears under a GRPO RL update, so the effect is a property of post-training rather than of one algorithm. It needs the safety data to come last, is selective across corpora, fades as the window becomes a smaller fraction of pretraining, and reproduces on a different model. More broadly, this is path dependence in post-training: the training path leaves an imprint that post-SFT behavior does not reveal. Because what comes last shapes the response to later alignment, what a model was trained on last should be reported along with its scores, so that downstream developers that align it next know what they are starting from.

\bibliography{references}

\clearpage
\appendix
\label{app:start} 

\section{Data, Metrics, and Sanity Checks}

All three-seed results use seeds 42, 43, and 44 for the SFT, DPO, and evaluation path. The SFT subset and DPO preference pairs are drawn from training splits, held out UltraFeedback preference pairs are reserved for the DPO proxy diagnostic in Table~\ref{tab:appendix-proxy}. We keep the same generation parameters and refusal classification path across branches. This makes branch to branch differences comparable under the controlled setup. 

The capability numbers of early OLMo fork are only used as sanity checks. Table~\ref{tab:branch-sanity} reports the post-SFT all branch capability sanity check used to verify that the comparison is not driven by a collapsed branch.

\begin{table}[!htbp]
\centering
\footnotesize
\setlength{\tabcolsep}{1.5pt}
\begin{tabular}{lcccccc}
\hline
Task & $C_{\mathrm{web}}$ & $C_{\mathrm{dclm}}$ & $C_{\mathrm{norm}}$ & $C_{\mathrm{safety}}$ & $C_{\mathrm{math}}$ & $C_{\mathrm{synth}}$ \\
\hline
MMLU & \shortstack{$25.6$\\$\pm0.1$} & \shortstack{$26.0$\\$\pm0.1$} & \shortstack{$25.5$\\$\pm0.1$} & \shortstack{$24.8$\\$\pm0.4$} & \shortstack{$25.5$\\$\pm0.1$} & \shortstack{$25.3$\\$\pm0.3$} \\[3pt]
HS acc-norm & \shortstack{$52.3$\\$\pm0.1$} & \shortstack{$52.2$\\$\pm0.1$} & \shortstack{$51.4$\\$\pm0.0$} & \shortstack{$51.4$\\$\pm0.1$} & \shortstack{$51.5$\\$\pm0.0$} & \shortstack{$51.8$\\$\pm0.0$} \\[3pt]
ARC-C acc-norm & \shortstack{$29.5$\\$\pm0.2$} & \shortstack{$29.6$\\$\pm0.2$} & \shortstack{$28.7$\\$\pm0.2$} & \shortstack{$28.7$\\$\pm0.4$} & \shortstack{$28.9$\\$\pm0.2$} & \shortstack{$29.7$\\$\pm0.3$} \\[3pt]
WinoGrande & \shortstack{$56.2$\\$\pm0.2$} & \shortstack{$56.8$\\$\pm0.2$} & \shortstack{$56.9$\\$\pm0.2$} & \shortstack{$55.2$\\$\pm0.6$} & \shortstack{$55.6$\\$\pm0.1$} & \shortstack{$57.1$\\$\pm0.1$} \\[3pt]
\hline
\end{tabular}
\caption{Post-SFT capability sanity checks for all 49B fork branches. Values are three-seed means in percentage with seed standard deviations, higher is better. HS is HellaSwag, ARC-C is ARC-Challenge, and acc-norm is the length normalized multiple choice score. Differences are small across branches.}
\label{tab:branch-sanity}
\end{table}

Table~\ref{tab:hparams} summarizes the run level configuration for each stage of the pipeline.

\begin{table}[!htbp]
\centering
\footnotesize
\setlength{\tabcolsep}{4pt}
\begin{tabular}{@{}p{0.13\columnwidth}p{0.27\columnwidth}p{0.52\columnwidth}@{}}
\hline
Stage & Data & Key settings \\
\hline
CPT & final window corpus & 500M tokens, lr $7.45{\times}10^{-5}$ \\
SFT & Tulu-style subset & 100k examples, lr $2{\times}10^{-5}$ \\
DPO & UltraFeedback pairs & 60k pairs, lr $5{\times}10^{-7}$, $\beta=0.1$ \\
Eval & refusal benchmarks & greedy decoding, 64 new tokens, batch 8 \\
\hline
\end{tabular}
\caption{Run level details for the main pipeline. CPT is continued pretraining. The DPO reference model is the corresponding SFT checkpoint.}
\label{tab:hparams}
\end{table}

\paragraph{Computing infrastructure.}
All experiments use 1B parameter models ($C_{\mathrm{web}}$--$C_{\mathrm{synth}}$ from OLMo-2-0425-1B and the Pythia-1B-deduped replication). Continued pretraining, SFT, and DPO run on NVIDIA A100 80GB GPUs (two GPUs with fully sharded data parallel for training, one GPU for generation based refusal evaluation) on a Linux cluster, inside an NGC PyTorch container (\texttt{nvcr.io/nvidia/pytorch}, 25.x). Training uses PyTorch with HuggingFace Transformers, TRL for SFT and DPO, and Accelerate with FSDP, capability metrics use the EleutherAI lm-evaluation-harness, and the refusal detector validation uses the WildGuard classifier. All post-training and evaluation use the OLMo-2-0425-1B-Instruct tokenizer and chat template. Exact library versions are pinned in the accompanying code release.

\paragraph{Refusal Detector Validation against WildGuard}
\label{app:wildguard}
We cross-validate the lexical detector against WildGuard~\citep{han2024wildguard}, a trained classifier that judges refusal from response semantics rather than surface strings, by regenerating completions for all $C_{\mathrm{web}}$/$C_{\mathrm{safety}}$ SFT and DPO checkpoints (3 seeds, $256$ new tokens) on five harmful refusal benchmarks and scoring each with both detectors on the same text. Table~\ref{tab:wildguard} reports the per benchmark $C_{\mathrm{web}}$--$C_{\mathrm{safety}}$ protection under both detectors, which never differ by more than $1.7$ points or change sign. Because this is a separate generation pass, absolute values differ slightly from Table~\ref{tab:main-erosion}, but the comparison is on identical completions.

\begin{table}[!htbp]
\centering
\footnotesize
\begin{tabular}{lcc}
\hline
Benchmark & Keyword $P$ & WildGuard $P$ \\
\hline
AdvBench          & $+3.1$  & $+1.4$  \\
XSTest-unsafe     & $+13.2$ & $+12.2$ \\
BeaverTails-clean & $+5.6$  & $+4.9$  \\
OR-Bench-Hard     & $+13.5$ & $+12.8$ \\
OR-Bench-Toxic    & $+11.3$ & $+9.8$  \\
\hline
Mean              & $+9.3$  & $+8.2$  \\
\hline
\end{tabular}
\caption{Refusal detector validation. $C_{\mathrm{web}}$--$C_{\mathrm{safety}}$ protection (percentage points) under the
lexical keyword detector versus WildGuard, on the same regenerated completions. The
protection reproduces under the non lexical classifier on every benchmark.}
\label{tab:wildguard}
\end{table}

\paragraph{Post-Training Sanity Checks}
These sanity checks confirm that $C_{\mathrm{safety}}$ remains a functional model after DPO. Table~\ref{tab:appendix-proxy} reports two sanity checks for the $C_{\mathrm{web}}$--$C_{\mathrm{safety}}$ contrast. First, IFEval~\citep{zhou2023instructionfollowing} does not show an instruction following collapse for $C_{\mathrm{safety}}$ after DPO. Prompt level scores are also matched across branches, while instruction level scores are slightly higher for $C_{\mathrm{safety}}$ but have larger seed variance. Second, on held out UltraFeedback preference pairs, $C_{\mathrm{safety}}$ has comparable DPO preference accuracy and a slightly larger chosen minus rejected margin gain, while its top-$32$ Kullback--Leibler (KL) proxy and mean absolute log-probability drift are lower. The results suggest that the refusal result is not caused by DPO failing to move the $C_{\mathrm{safety}}$ model, but it does not identify a mechanism.

\begin{table}[!htbp]
\centering
\small
\begin{tabular}{lcc}
\hline
Diagnostic & $C_{\mathrm{web}}$ & $C_{\mathrm{safety}}$ \\
\hline
IFEval prompt strict (\%) & $17.0\pm0.3$ & $16.8\pm1.8$ \\
IFEval instruction strict (\%) & $28.2\pm0.3$ & $28.8\pm1.3$ \\
Preference accuracy (\%) & $49.3\pm0.6$ & $50.0\pm0.4$ \\
Margin gain ($\times 100$) & $1.81\pm0.10$ & $2.17\pm0.11$ \\
Abs. log-prob. drift ($\times 100$) & $10.58\pm0.13$ & $9.87\pm0.13$ \\
Top-$32$ KL proxy ($\times 100$) & $1.62\pm0.07$ & $1.46\pm0.06$ \\
\hline
\end{tabular}
\caption{Instruction following and held out DPO proxy checks for the $C_{\mathrm{web}}$--$C_{\mathrm{safety}}$ contrast. Values are three-seed mean $\pm$ standard deviation. The KL proxy is computed tokenwise over the union of the SFT and DPO top-$32$ tokens on held out UltraFeedback completions, so it is a relative drift diagnostic rather than a full vocabulary KL. Lower drift with comparable margin gain is favorable, but we use this as supporting evidence rather than a mechanism claim.}
\label{tab:appendix-proxy}
\end{table}

Table~\ref{tab:appendix-utility} reports a separate post-DPO capability sanity check. The tasks measure general multiple choice capability after the same DPO run. The safety branch is close on MMLU, HellaSwag, and ARC-Challenge, but lower on WinoGrande, giving a four task mean cost of $0.9$ percentage points across three seeds. 

\begin{table}[!htbp]
\centering
\small
\setlength{\tabcolsep}{4pt}
\begin{tabular}{lcc}
\hline
Diagnostic & $C_{\mathrm{web}}$ & $C_{\mathrm{safety}}$ \\
\hline
MMLU acc. (\%) & $25.8\pm0.2$ & $25.2\pm0.3$ \\
HellaSwag acc-norm (\%) & $52.4\pm0.1$ & $51.6\pm0.1$ \\
ARC-C acc-norm (\%) & $29.7\pm0.2$ & $28.9\pm0.6$ \\
WinoGrande acc. (\%) & $56.3\pm0.2$ & $55.0\pm0.8$ \\
Four task mean (\%) & $41.1\pm0.1$ & $40.2\pm0.2$ \\
\hline
\end{tabular}
\caption{Post-DPO capability sanity check for the $C_{\mathrm{web}}$--$C_{\mathrm{safety}}$ contrast. Values are three-seed mean $\pm$ standard deviation from lm-evaluation-harness~\citep{gao2021framework} on the final DPO checkpoints. ARC-C is ARC-Challenge, acc-norm is length normalized accuracy. Higher is better. The safety branch shows a small capability cost, mainly on WinoGrande, so the refusal result should be read together with utility and over-refusal checks.}
\label{tab:appendix-utility}
\end{table}

\begin{figure}[!htbp]
\centering
\includegraphics[width=.86\linewidth]{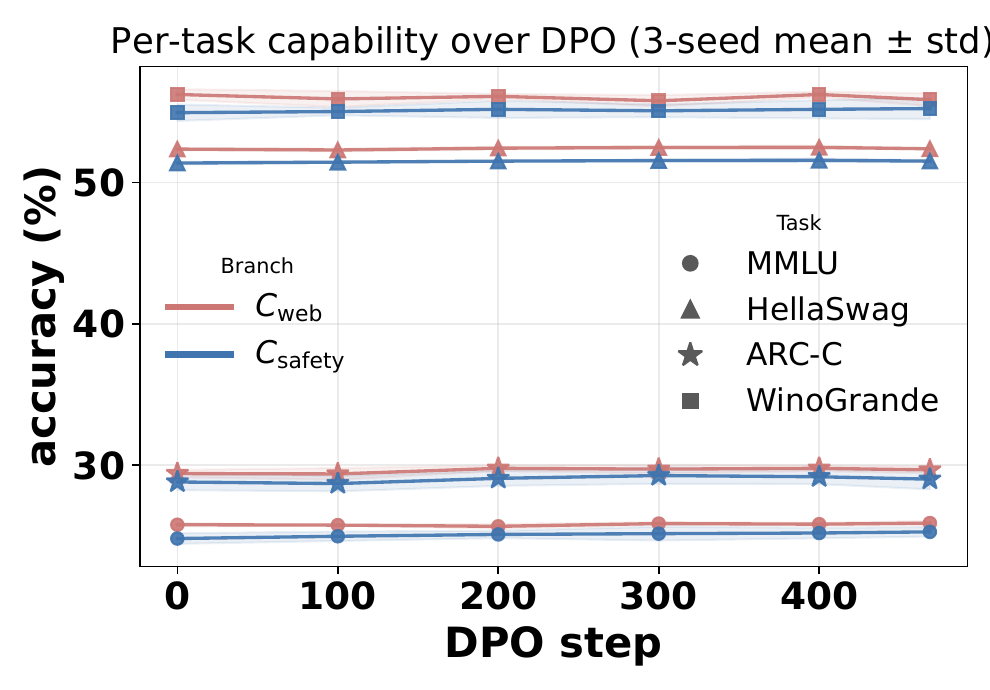}
\caption{Per task capability over DPO in a combined view. Color marks the branch ($C_{\mathrm{web}}$ vs $C_{\mathrm{safety}}$) and marker shape marks the task. Values are three-seed means with seed standard deviation bands, and higher is better.}
\label{fig:utility-pertask-combined}
\end{figure}

To check whether this capability cost is created by DPO, we also track per task capability across DPO checkpoints (Figure~\ref{fig:utility-pertask-combined}), the utility side counterpart to the refusal erosion in Figure~\ref{fig:erosion-curve}. Because SFT and DPO are identical across branches, this gap is attributable to the final window pretraining rather than to either post-training stage, so the safety branch carries a small capability cost from pretraining but pays almost nothing further during DPO. The final step values reproduce Table~\ref{tab:appendix-utility} within seed noise, a consistency check on the re-run.

\section{Decontamination and Repetition}
\label{app:decontam}

We audit prompt overlap between the harmful refusal evaluations and the $C_{\mathrm{safety}}$ source data. AdvBench and XSTest-unsafe have low overlap: 2/520 prompts for AdvBench and 3/200 prompts for XSTest-unsafe. BeaverTails has substantial overlap, 196/1000 prompts, including 185 exact matches. For this reason, the results in Table~\ref{tab:main-erosion} are reported on the decontaminated prompt sets, with AdvBench and XSTest unaffected and BeaverTails restricted to its non-overlapping prompts.

\begin{figure}[!htbp]
\centering
\includegraphics[width=\linewidth]{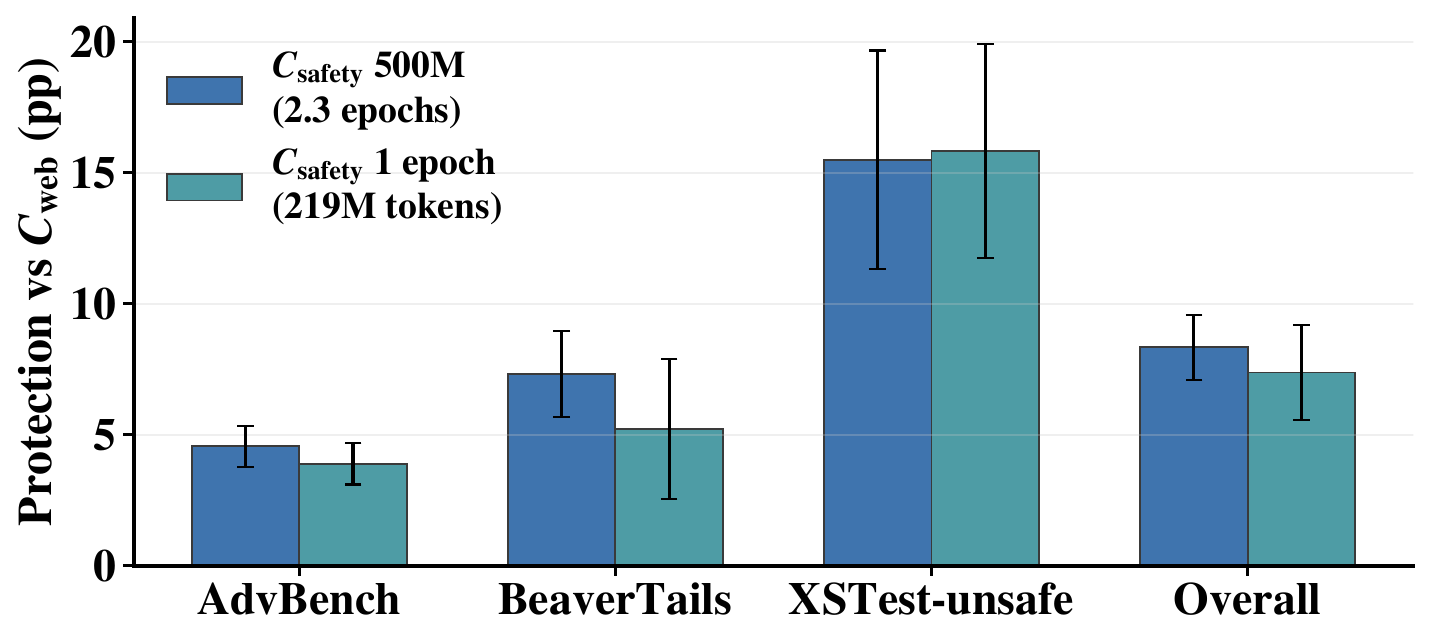}
\caption{Repetition control for the safety corpus. Lower erosion is better. One unique pass recovers most protection on the clean headline benchmarks.}
\label{fig:repetition}
\end{figure}

A further potential confound is repetition rather than content. The full $C_{\mathrm{safety}}$ branch repeats a 219M token corpus to reach the 500M token budget. We therefore reconstruct a $C_{\mathrm{safety}}$ checkpoint with one single pass and rerun the same SFT/DPO/evaluation path with three seeds. Table~\ref{tab:repetition} and Figure~\ref{fig:repetition} show that one unique pass recovers most of the full dose effect on the clean headline benchmarks. 

\begin{table}[!htbp]
\centering
\footnotesize
\setlength{\tabcolsep}{4pt}
\begin{tabular}{lccc}
\hline
Benchmark & \shortstack{$C_{\mathrm{web}}$\\(500M)} & \shortstack{full $C_{\mathrm{safety}}$\\(500M)} & \shortstack{one pass\\(219M)} \\
\hline
AdvBench & $11.8\pm0.4$ & $7.2\pm0.4$ & $7.9\pm0.7$ \\
BeaverTails & $35.3\pm1.2$ & $28.0\pm0.5$ & $30.1\pm3.2$ \\
XSTest-unsafe & $51.2\pm1.4$ & $35.7\pm0.6$ & $35.3\pm1.4$ \\
\hline
\end{tabular}
\caption{Repetition control for $C_{\mathrm{safety}}$. Values are three-seed mean DPO erosion in percentage points with standard deviations, from an independent re-run, so the $C_{\mathrm{web}}$ and full $C_{\mathrm{safety}}$ columns differ slightly from Table~\ref{tab:main-erosion}. Lower is less erosion. The one pass safety corpus matches the full dose result closely on XSTest-unsafe and keeps the same direction on AdvBench.}
\label{tab:repetition}
\end{table}

\begin{figure*}[!htbp]
\centering
\includegraphics[width=\textwidth]{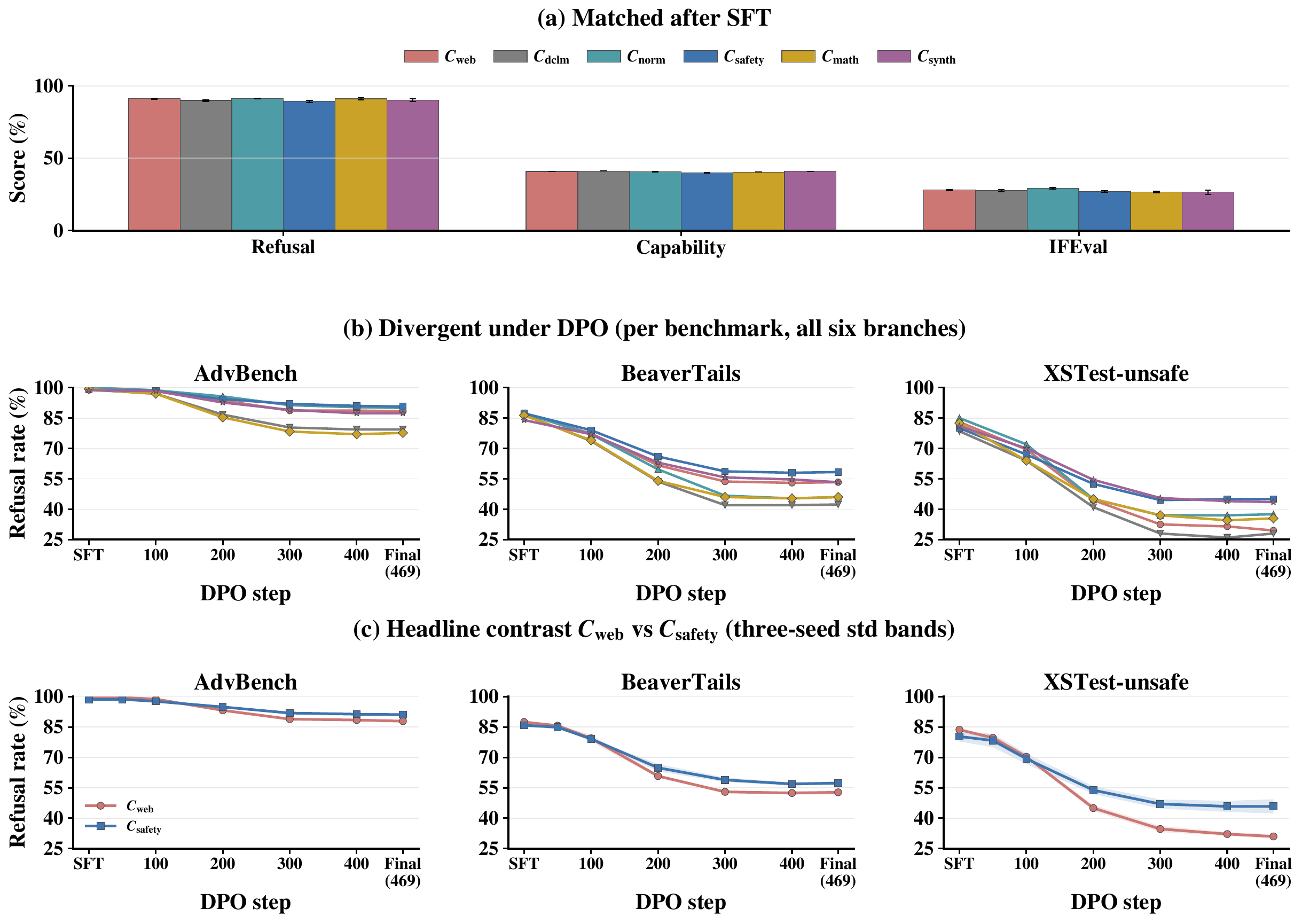}
\caption{All branches, matched after SFT and divergent under DPO. \textbf{(a)} At the SFT checkpoint all six branches are closely matched on the three release criteria, refusal (overall over AdvBench, BeaverTails, and XSTest-unsafe), capability (mean accuracy over MMLU, HellaSwag, ARC-Challenge, and WinoGrande), and instruction following (IFEval instruction level strict), each within a few points. Error bars are standard deviations over three SFT seeds. \textbf{(b)} Under the same DPO recipe their refusal diverges on AdvBench, BeaverTails, and XSTest-unsafe (single-seed DPO trajectory for cross-branch comparison). The non-safety branches show benchmark specific effects, while $C_{\mathrm{safety}}$ is the most consistent harmful request refusal branch. \textbf{(c)} The headline $C_{\mathrm{web}}$ versus $C_{\mathrm{safety}}$ trajectories with three-seed standard deviation bands, on the same dense checkpoints as Figure~\ref{fig:erosion-curve}(b), for context on the seed variance of the row above.}
\label{fig:all-dpo-curves}
\end{figure*}

\section{All Branch Refusal Views}

The following results collect branch wide refusal views that are summarized in the main text. Figure~\ref{fig:all-dpo-curves} shows that all six branches are closely matched after SFT and then diverge under DPO, with $C_{\mathrm{safety}}$ the most consistent harmful request refusal branch and $C_{\mathrm{synth}}$ only a partial positive branch.

\paragraph{Significance of the divergence.} Welch's t-tests on overall refusal over three seeds confirm the pattern. After SFT no branch differs significantly from $C_{\mathrm{safety}}$ (all $p>0.05$, Table~\ref{tab:allbranch-ttest}). After DPO, $C_{\mathrm{safety}}$ refuses significantly more than $C_{\mathrm{web}}$, $C_{\mathrm{dclm}}$, $C_{\mathrm{norm}}$, and $C_{\mathrm{math}}$ (all $p<0.02$), while $C_{\mathrm{synth}}$ remains not significant ($p=0.59$), matching its status as the one partial positive branch. A one way ANOVA across the six branches sharpens from $F=3.3$ at SFT to $F=16.4$ ($p=5\times10^{-5}$) under DPO.

\begin{table}[!htbp]
\centering
\small
\setlength{\tabcolsep}{6pt}
\begin{tabular}{lcc}
\hline
Branch vs $C_{\mathrm{safety}}$ & After SFT ($p$) & After DPO ($p$) \\
\hline
$C_{\mathrm{web}}$ & $0.069$ & $0.002$ \\
$C_{\mathrm{dclm}}$ & $0.424$ & $0.014$ \\
$C_{\mathrm{norm}}$ & $0.071$ & $0.018$ \\
$C_{\mathrm{math}}$ & $0.083$ & $0.012$ \\
$C_{\mathrm{synth}}$ & $0.456$ & $0.590$ \\
\hline
\end{tabular}
\caption{Welch's t-test $p$-values for overall refusal (three seeds) comparing each branch to $C_{\mathrm{safety}}$, before (SFT) and after (DPO) preference optimization. No branch differs from $C_{\mathrm{safety}}$ after SFT. After DPO every branch except $C_{\mathrm{synth}}$ separates from $C_{\mathrm{safety}}$, consistent with $C_{\mathrm{synth}}$ being the one partial positive branch.}
\label{tab:allbranch-ttest}
\end{table}

\section{Boundary Checks}

The saturated base boundary check repeats the comparison from a later OLMo-2-1B checkpoint at roughly 4T pretraining tokens. The same absolute 500M token intervention is much smaller relative to the prior training history and does not reproduce the early fork $C_{\mathrm{safety}}$ effect. Near the saturated base the branches sit close to ceiling on AdvBench, so erosion is uninformative there and Table~\ref{tab:boundary} compares post-DPO refusal endpoints directly, the $C_{\mathrm{safety}}$--$C_{\mathrm{web}}$ gap has vanished, with $C_{\mathrm{safety}}$ no higher than $C_{\mathrm{web}}$ on any benchmark, consistent with overall protection falling to about $0.5$ point here from $9.1$ at the 49B fork. We interpret this as a relative dose boundary, not as evidence that late data cannot move trained models.

\begin{table}[!htbp]
\centering
\small
\begin{tabular}{lccc}
\hline
Branch & Adv. & Beaver & XSTest \\
\hline
$C_{\mathrm{web}}$~(4T) & $99.7\pm0.0$ & $70.9\pm1.6$ & $84.7\pm2.2$ \\
$C_{\mathrm{safety}}$~(4T) & $99.9\pm0.2$ & $69.9\pm3.2$ & $80.7\pm2.1$ \\
$C_{\mathrm{safety}}{-}C_{\mathrm{web}}$ & $+0.2$ & $-1.0$ & $-4.0$ \\
\hline
\end{tabular}
\caption{Saturated base boundary check. Values are three-seed post-DPO refusal percentages with seed standard deviations, the last row is the $C_{\mathrm{safety}}$~(4T) minus $C_{\mathrm{web}}$~(4T) endpoint gap in percentage points. Higher refusal is better on these harmful request benchmarks. The saturated base endpoint gap does not reproduce the 49B fork effect.}
\label{tab:boundary}
\end{table}

\section{Baseline without Continued Pretraining}
\label{app:base-control}

To separate the content of the final window from the act of adding one, we run the same SFT and DPO recipe directly on the 49B token fork with no 500M token continued pretraining. This Base branch shares the fork, the SFT data, and the DPO data with $C_{\mathrm{web}}$ and $C_{\mathrm{safety}}$, so it isolates the act of continued pretraining from its content. Because all three branches receive the same safety containing SFT mixture, Base starts from a comparable SFT refusal level (overall $87.6$, against $89.5$ for $C_{\mathrm{web}}$ and $88.2$ for $C_{\mathrm{safety}}$), so the comparison is again about erosion under DPO rather than the starting point. Table~\ref{tab:base-control} reports decontaminated three-seed erosion on the same basis as Table~\ref{tab:main-erosion}. First, the headline contrast does not come from continuation versus none. Both $C_{\mathrm{web}}$ and $C_{\mathrm{safety}}$ continue for 500M tokens, and safety content still loses about $8$ points less refusal than web content. Second, continuation itself is mildly stabilizing. The Base branch erodes more than $C_{\mathrm{web}}$ on every benchmark, about $5.6$ points more overall, and safety content adds a further reduction, for a total Base to $C_{\mathrm{safety}}$ gap of about $13.8$ points overall. The controlled $C_{\mathrm{web}}$--$C_{\mathrm{safety}}$ comparison is therefore conservative. It measures the effect of the final window content on top of a continuation that already reduces erosion a little on its own.

\begin{table}[!htbp]
\centering
\small
\begin{tabular}{lccc}
\hline
Benchmark & $E$(Base) & $E(C_{\mathrm{web}})$ & $E(C_{\mathrm{safety}})$ \\
\hline
AdvBench & $21.6\pm1.5$ & $11.7\pm0.3$ & $8.4\pm0.2$ \\
BeaverTails & $36.7\pm0.8$ & $33.0\pm1.5$ & $27.8\pm0.6$ \\
XSTest-unsafe & $55.0\pm3.1$ & $51.9\pm1.2$ & $35.7\pm0.6$ \\
\hline
Overall & $37.8\pm0.7$ & $32.2\pm0.8$ & $24.0\pm0.1$ \\
\hline
\end{tabular}
\caption{Results with no continuation baseline. Base is the 49B token fork with SFT and DPO applied directly, with no 500M final window, under the same recipe as $C_{\mathrm{web}}$ and $C_{\mathrm{safety}}$. Values are three-seed mean DPO erosion $E_b(c)=R_b(\theta_c^S)-R_b(\theta_c^D)$ in percentage points on prompts decontaminated against the $C_{\mathrm{safety}}$ source data, the same basis as Table~\ref{tab:main-erosion}. Lower erosion is better. Erosion increases from $C_{\mathrm{safety}}$ to $C_{\mathrm{web}}$ to Base on every benchmark, so the final window content effect sits on top of a mild protection of continuation itself.}
\label{tab:base-control}
\end{table}

\section{Mechanism Diagnostics}

We also checked whether the safety branch moves less during DPO. Table~\ref{tab:dpo-movement} reports a single seed $C_{\mathrm{web}}$-vs-$C_{\mathrm{safety}}$ diagnostic: the mean relative parameter update norm is about $0.0017$ per layer for both branches at step 200 and at the final DPO checkpoint. Hidden state drift is also similar in magnitude, with small layerwise differences that are not stable enough to interpret mechanistically. A refusal direction projection probe~\citep{arditi2024refusal} did not cleanly separate $C_{\mathrm{web}}$ and $C_{\mathrm{safety}}$. The diagnostic only rules out the trivial explanation that $C_{\mathrm{safety}}$ preserves refusal because DPO barely changes it.

\begin{table}[!htbp]
\centering
\small
\setlength{\tabcolsep}{4pt}
\begin{tabular}{lcc}
\hline
Diagnostic & $C_{\mathrm{web}}$ & $C_{\mathrm{safety}}$ \\
\hline
Rel. update, step 200 ($\times10^{-3}$) & $1.656$ & $1.656$ \\
Rel. update, final ($\times10^{-3}$) & $1.667$ & $1.668$ \\
Harmful drift, final ($\times10^{-2}$) & $1.87$ & $1.92$ \\
Benign drift, final ($\times10^{-2}$) & $1.56$ & $1.57$ \\
Borderline drift, final ($\times10^{-2}$) & $2.01$ & $2.24$ \\
\hline
\end{tabular}
\caption{DPO movement diagnostic for a single $C_{\mathrm{web}}$-vs-$C_{\mathrm{safety}}$ seed. Relative update norms average transformer layer parameter changes normalized by SFT weight norms. Hidden state drift averages cosine drift across non-embedding layers and prompts. Similar magnitudes rule out the simplest explanation that $C_{\mathrm{safety}}$ preserves refusal because DPO barely updates it.}
\label{tab:dpo-movement}
\end{table}

\section{Check on Pythia}

We also ran a fuller check on Pythia-1B-deduped at the 50B token region~\citep{biderman2023pythia}. Starting from checkpoint step 24{,}000, we applied the same 500M token $C_{\mathrm{web}}$ and $C_{\mathrm{safety}}$ final window continued pretraining interventions, then ran the same SFT, DPO, and refusal evaluation pipeline with three seeds for SFT and DPO.

\begin{figure*}[t]
\centering
\includegraphics[width=.92\textwidth]{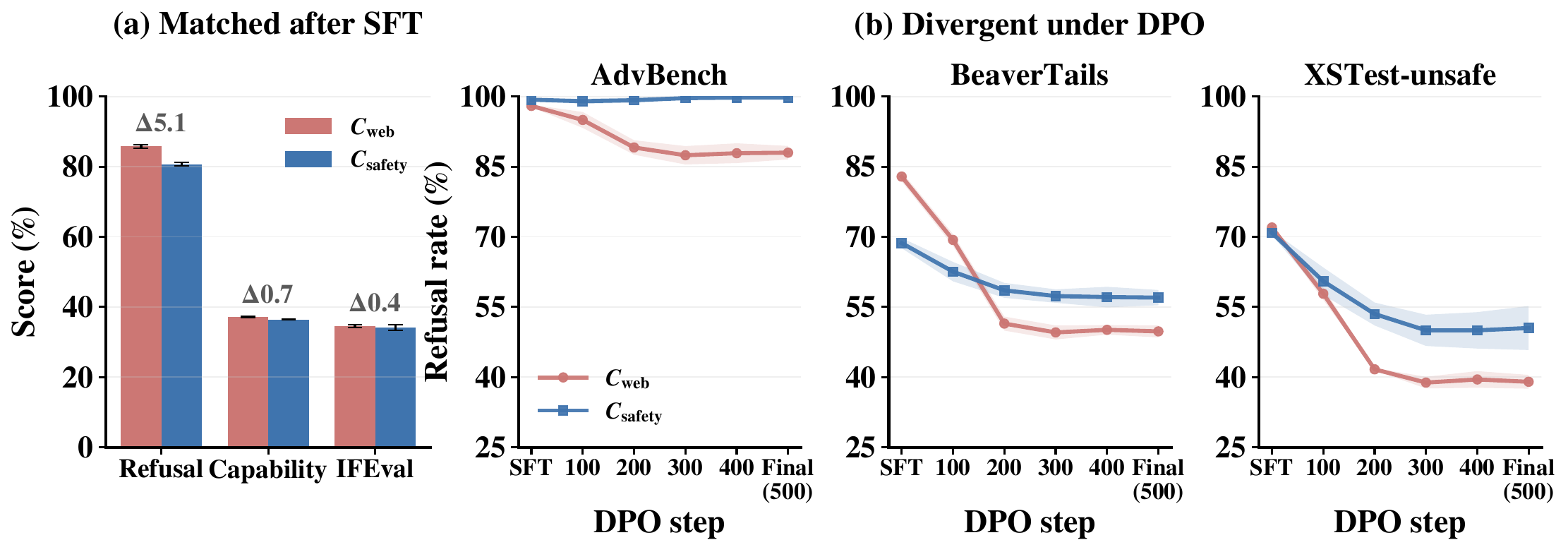}
\caption{Pythia replication of the matched then divergent pattern for the $C_{\mathrm{web}}$ versus $C_{\mathrm{safety}}$ contrast, on Pythia-1B-deduped forked at checkpoint step 24{,}000 (about 50B tokens). \textbf{(a)} At the SFT checkpoint the two Pythia branches are closely matched on the three release criteria, refusal (overall over AdvBench, BeaverTails, and XSTest-unsafe), capability (mean accuracy over MMLU, HellaSwag, ARC-Challenge, and WinoGrande), and instruction following (IFEval instruction level strict). \textbf{(b)} Under the same DPO recipe their refusal on AdvBench, BeaverTails, and XSTest-unsafe diverges, and the safety branch loses less refusal even though its SFT starting point is not uniformly higher than the web branch. Higher refusal is better. Error bars and shaded bands are standard deviations over three SFT and DPO seeds.}
\label{fig:pythia-dense}
\end{figure*}

The Pythia result supports the direction of the main finding (Figure~\ref{fig:pythia-dense}, Table~\ref{tab:pythia-external}), with lower $C_{\mathrm{safety}}$ erosion on all three refusal benchmarks. The SFT starting point is slightly different across branches: $C_{\mathrm{safety}}$ starts lower overall than $C_{\mathrm{web}}$ ($79.6$ vs. $84.3$ refusal), mainly because BeaverTails is lower before DPO. After DPO, however, $C_{\mathrm{safety}}$ is higher overall ($69.1$ vs. $59.0$ refusal). OR-Bench-Hard over-refusal is matched in this check on Pythia ($37.8\pm1.4$ for $C_{\mathrm{web}}$ and $37.7\pm0.9$ for $C_{\mathrm{safety}}$).

\begin{table}[!htbp]
\centering
\small
\setlength{\tabcolsep}{4pt}
\begin{tabular}{lccc}
\hline
Benchmark & $E(C_{\mathrm{web}})$ & $E(C_{\mathrm{safety}})$ & $P(C_{\mathrm{safety}})$ \\
\hline
AdvBench & $10.0\pm1.8$ & $-0.3\pm0.0$ & $+10.3\pm1.8$ \\
BeaverTails & $32.7\pm2.6$ & $11.6\pm0.3$ & $+21.1\pm2.8$ \\
XSTest-unsafe & $33.2\pm1.9$ & $20.2\pm4.0$ & $+13.0\pm2.9$ \\
\hline
Overall & $25.3\pm1.6$ & $10.5\pm1.4$ & $+14.8\pm1.5$ \\
\hline
\end{tabular}
\caption{DPO refusal erosion on Pythia-1B-deduped at the 50B token region, the same $C_{\mathrm{web}}$/$C_{\mathrm{safety}}$ final window contrast on a different model. For benchmark $b$, $E_b(c)=R_b(\theta_c^S)-R_b(\theta_c^D)$ is the refusal lost from SFT to DPO and $P_b(C_{\mathrm{safety}})=E_b(C_{\mathrm{web}})-E_b(C_{\mathrm{safety}})$ is protection relative to $C_{\mathrm{web}}$. Values are three-seed means in percentage points with seed standard deviations, paired across seeds for protection. Lower erosion and higher protection are better: $C_{\mathrm{safety}}$ loses less refusal on every benchmark, and Overall is the mean of the three benchmarks. SFT and DPO endpoints are given in the text.}
\label{tab:pythia-external}
\end{table}

\section{Robustness Across Post-Training Recipes}
The $C_{\mathrm{web}}$--$C_{\mathrm{safety}}$ divergence persists when we swap components of the post-training recipe. We report two such variations: a second RL task under GRPO, and DPO with a different preference source.

\begin{figure}[!htbp]
\centering
\includegraphics[width=\linewidth]{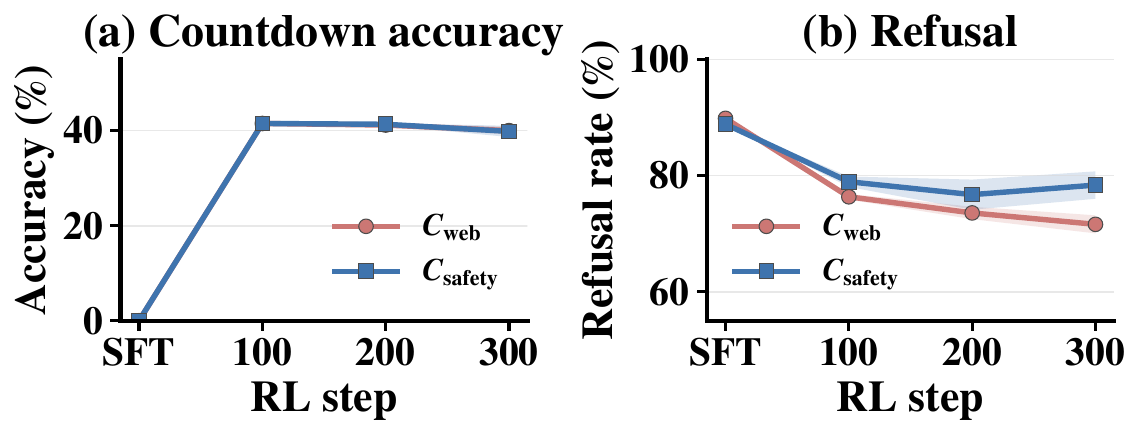}
\caption{The same math reinforcement learning (GRPO on Countdown) from the $C_{\mathrm{web}}$ and $C_{\mathrm{safety}}$ SFT checkpoints. \textbf{(a)} Both branches learn Countdown equally, their accuracy trajectories overlap. \textbf{(b)} Under the same update their overall refusal diverges and $C_{\mathrm{safety}}$ loses far less. Refusal is the mean over AdvBench, BeaverTails, and XSTest-unsafe. Lines are three-seed means and bands are seed standard deviations.}
\label{fig:rl-countdown}
\end{figure}

\paragraph{RL task.}\label{app:rl-countdown}
Beyond the GCD task in the main text (Section~\ref{sec:not-dpo}), we test a second RL task. We apply the same GRPO update with a verifiable reward on Countdown, a procedurally generated arithmetic task in which the model must combine given numbers into a target. Each instance draws three or four distinct integers from $1$ to $20$ together with a target from $1$ to $100$ that is reachable through the four basic operations, and we generate synthetic data for this task. We produce $4000$ training and $400$ evaluation instances, deduplicated by their numbers and target so that the two sets are disjoint, and Countdown accuracy is measured on the held out evaluation set. Starting from the $C_{\mathrm{web}}$ and $C_{\mathrm{safety}}$ SFT checkpoints and running the identical recipe over three seeds, both branches learn the task equally, reaching $40.0\pm0.8$ and $39.8\pm1.2$ percent accuracy along overlapping trajectories (Figure~\ref{fig:rl-countdown}a). Because trainability is matched, a difference in refusal cannot be attributed to how much the update moves each branch. Yet refusal diverges under the same update (Figure~\ref{fig:rl-countdown}b): $C_{\mathrm{web}}$ loses $18.2\pm1.6$ points of overall refusal while $C_{\mathrm{safety}}$ loses $10.5\pm2.4$, a protection of $7.7\pm3.4$ points over three seeds, largest on XSTest-unsafe. The signature matches both the DPO result and the GCD task, with $C_{\mathrm{safety}}$ starting no higher after SFT yet ending well above $C_{\mathrm{web}}$. The same matched trainability with divergent refusal therefore appears on two different RL tasks, so the effect generalizes beyond preference optimization to RL rather than depending on one particular task.

\paragraph{Preference data.}\label{app:pref-robustness}
The main pipeline draws its DPO preferences from UltraFeedback~\citep{ultrafeedback2023}, scored by AI feedback. To test whether the protection is specific to that source, we repeat the DPO from the same SFT checkpoints with an independent human preference dataset, LMSYS Chatbot Arena~\citep{zheng2023judging}, whose labels are real pairwise votes from users of a public model arena rather than synthetic AI feedback. The recipe is identical and each branch uses three seeds. Both branches start from the same SFT checkpoints, and their refusal is matched at the start, if anything slightly higher for $C_{\mathrm{web}}$ (overall $89.5$ versus $88.2$ on decontaminated prompts, and $C_{\mathrm{web}}$ at least as high on every benchmark), so a smaller erosion for $C_{\mathrm{safety}}$ cannot come from a higher starting point. Table~\ref{tab:pref-robustness} reports the decontaminated per benchmark erosion and protection, on the same basis as Table~\ref{tab:main-erosion}. The protection reproduces: $C_{\mathrm{web}}$ loses more refusal than $C_{\mathrm{safety}}$ on every benchmark, with overall protection $2.3\pm0.4$ points and the largest gap on XSTest-unsafe. Chatbot Arena is also a much smaller preference set, about $20$k usable single-turn comparisons and one epoch of roughly $160$ DPO steps against the $60$k pairs and $469$ steps of UltraFeedback, so it erodes less total refusal and the protection magnitude is smaller. Even so, erosion is clearly present and the branch ordering is stable across seeds, with $C_{\mathrm{safety}}$ protected on every benchmark.

\begin{table}[!htbp]
\centering
\footnotesize
\setlength{\tabcolsep}{4pt}
\begin{tabular}{lccc}
\hline
Benchmark & $E_b(C_{\mathrm{web}})$ & $E_b(C_{\mathrm{safety}})$ & $P_b(C_{\mathrm{safety}})$ \\
\hline
AdvBench & $1.6\pm0.2$ & $1.0\pm0.2$ & $0.6\pm0.0$ \\
BeaverTails & $10.2\pm0.6$ & $7.7\pm0.4$ & $2.5\pm0.3$ \\
XSTest-unsafe & $15.5\pm0.6$ & $11.7\pm0.8$ & $3.8\pm1.3$ \\
\hline
Overall & $9.1\pm0.2$ & $6.8\pm0.5$ & $2.3\pm0.4$ \\
\hline
\end{tabular}
\caption{DPO refusal erosion under Chatbot Arena preferences~\citep{zheng2023judging} for the $C_{\mathrm{web}}$ and $C_{\mathrm{safety}}$ branches, from the same matched SFT checkpoints and on the same decontaminated basis as Table~\ref{tab:main-erosion}. For benchmark $b$, $E_b(c)=R_b(\theta_c^S)-R_b(\theta_c^D)$ is the refusal lost from SFT to DPO, and $P_b(C_{\mathrm{safety}})=E_b(C_{\mathrm{web}})-E_b(C_{\mathrm{safety}})$ is $C_{\mathrm{safety}}$'s protection relative to $C_{\mathrm{web}}$. Values are three-seed means in percentage points with seed standard deviations, the protection standard deviation paired across seeds. Positive protection means $C_{\mathrm{safety}}$ loses less refusal, reproducing the main result on an independent human preference dataset.}
\label{tab:pref-robustness}
\end{table}

Under a lower DPO learning rate ($2{\times}10^{-7}$, three seeds), the post-DPO endpoints are close and the OR-Bench-Hard over-refusal cost is also close ($50.4\pm1.3$ for $C_{\mathrm{web}}$ and $51.4\pm1.6$ for $C_{\mathrm{safety}}$).

\section{Corpus Diagnostics}
\label{app:corpus}

The diagnostic is a purely lexical probe of corpus content, not of model behavior. For each 49B fork final window corpus we sample $4{,}000$ documents and score each against five keyword families: assistant style refusal phrasing (mirroring the evaluation refusal markers), safety and policy vocabulary, harmful request topics, cautious hedging, and third person safety critique or draft review language. For every corpus and family we report the fraction of sampled documents that contain at least one case insensitive match, so the values are document hit rates rather than model outputs.

\begin{figure}[!htbp]
\centering
\includegraphics[width=\linewidth]{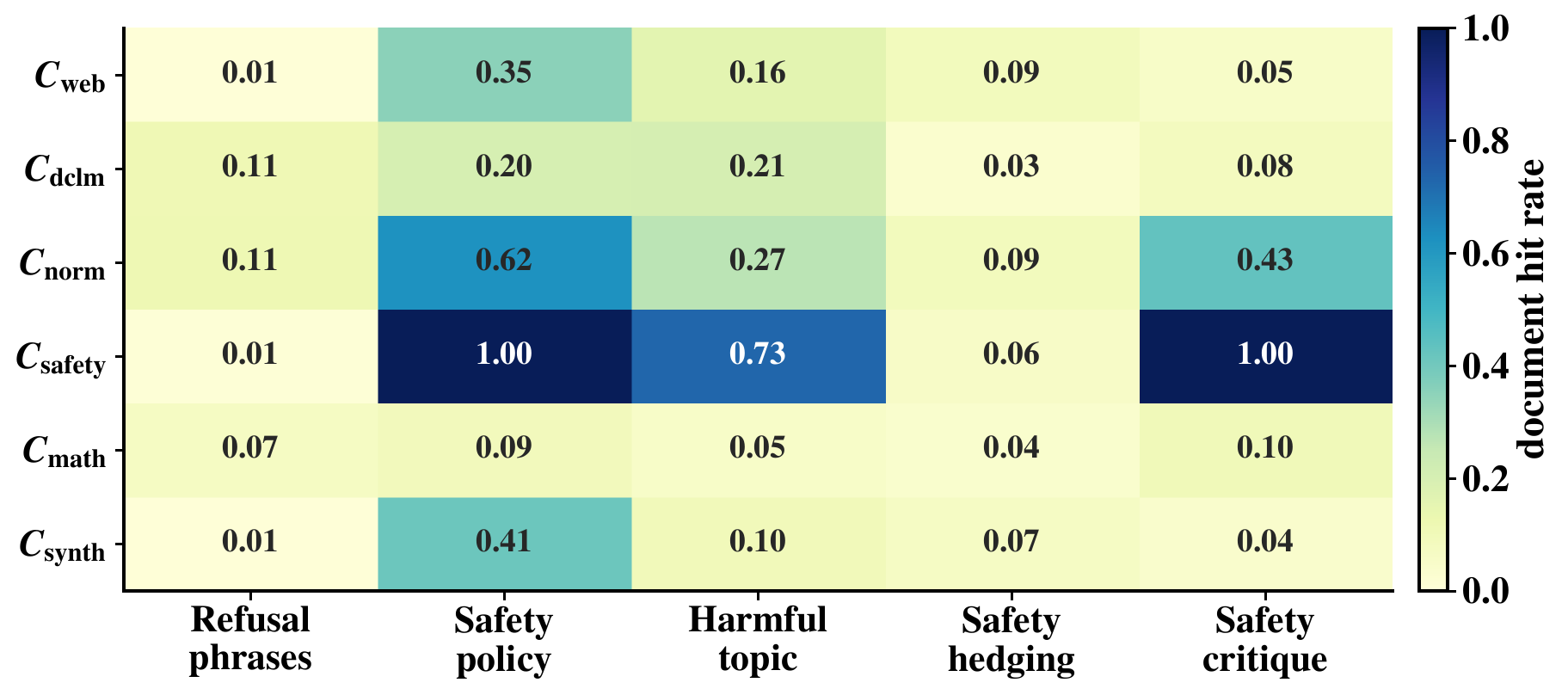}
\caption{Corpus diagnostics. Higher values mean more keyword hits for the named corpus feature, not necessarily better model behavior. The safety branch is characterized by harmful scenario grounding and safety critique rather than surface refusal templates alone. $C_{\mathrm{dclm}}$ is included as the DCLM contrast.}
\label{fig:corpus}
\end{figure}

Figure~\ref{fig:corpus} compares refusal template frequency, safety vocabulary, harmful topic density, hedging, and safety signals across all 49B fork corpora. It supports the hypothesis that harmful scenario grounding plus safety critique is relevant, while $C_{\mathrm{safety}}$ is not merely a refusal template corpus. The $C_{\mathrm{dclm}}$ branch is a useful contrast: it has nontrivial refusal template hits but much weaker safety policy and safety than $C_{\mathrm{safety}}$. These diagnostics remain correlational, but they are consistent with, though they do not establish, the conjecture that refusal grounded in safety reasoning and critique is harder for post-training to erode than surface refusal phrasing.

\section{Additional Probes}
\label{app:fragility}
\begin{figure}[h!]
\centering
\includegraphics[width=\linewidth]{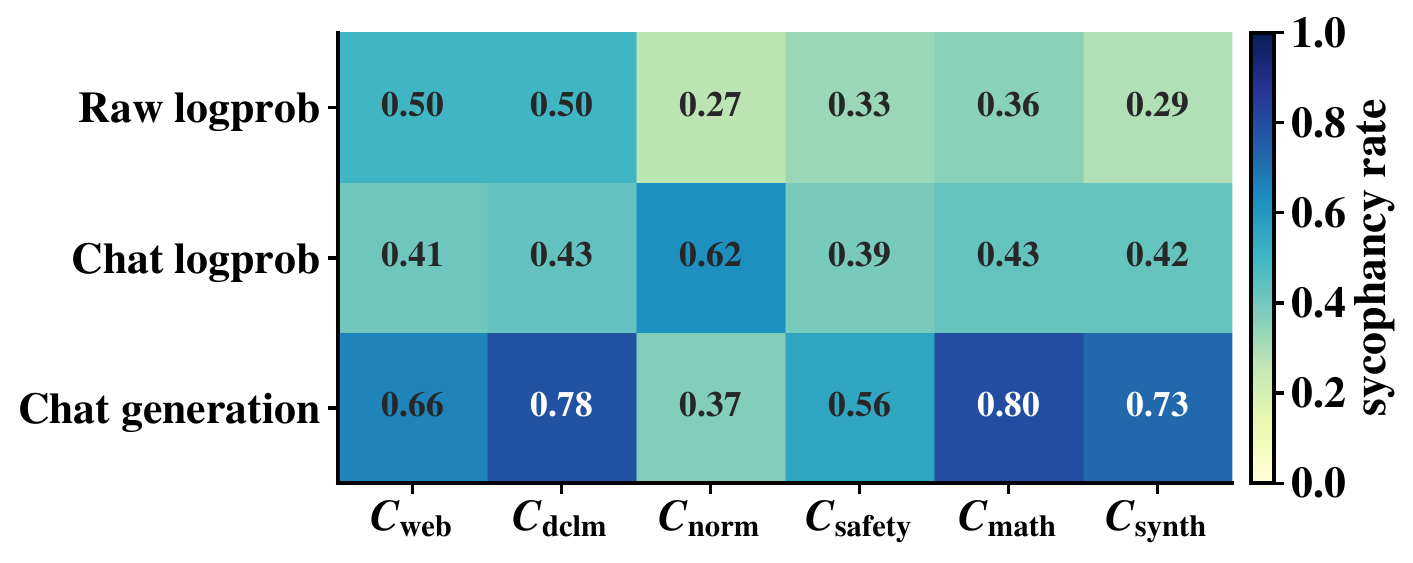}\\[2pt]
{\footnotesize (a) Sycophancy probe fragility}\\[6pt]
\includegraphics[width=\linewidth]{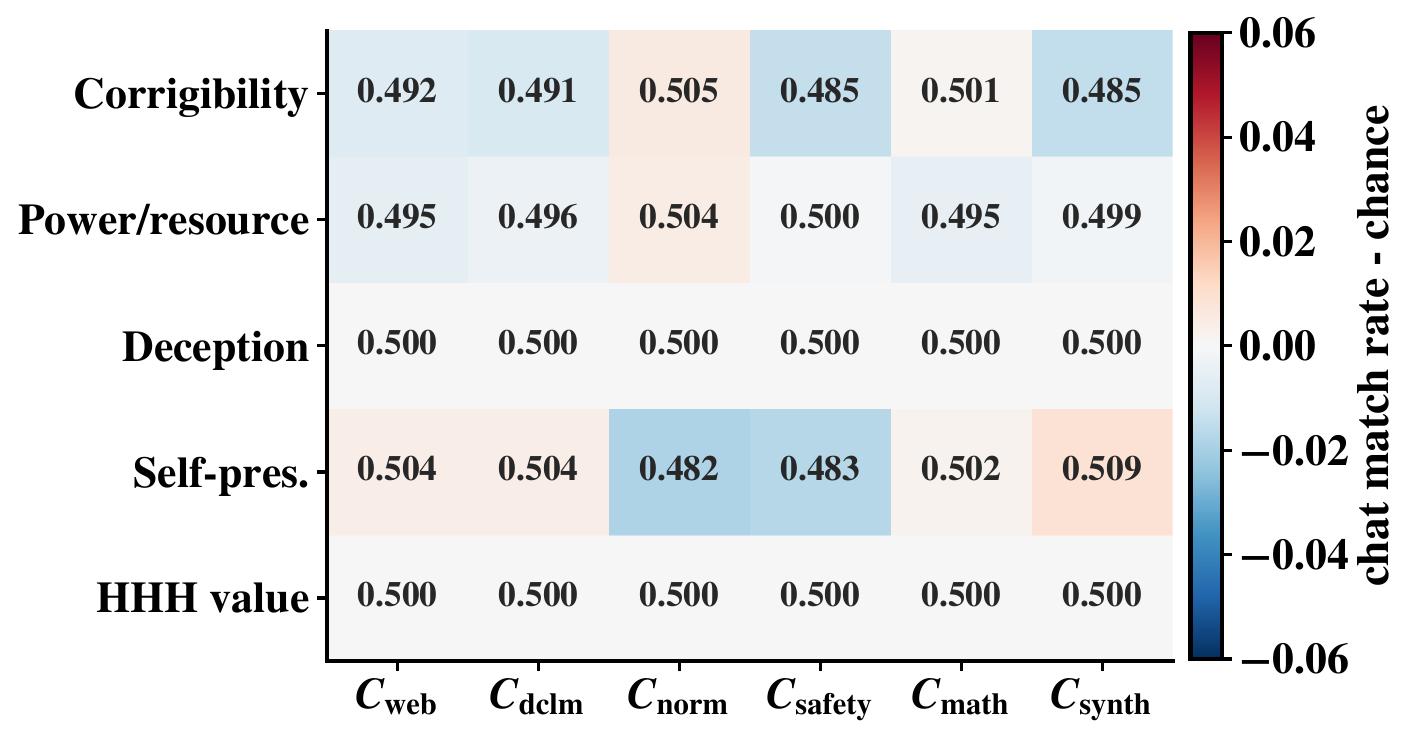}\\[2pt]
{\footnotesize (b) Persona and value probes}
\caption{Exploratory value probes, both null under a change of measurement. \textbf{(a)}~The sycophancy branch ranking changes across raw log-probability, chat log-probability, and chat generation scoring, so sycophancy is not used as evidence for a main claim. \textbf{(b)}~Cells show branch differences on the persona and value probes, there is no single higher is better direction across rows.}
\label{fig:probes}
\end{figure}

We ran two exploratory value probes as guardrails that keep the claim narrow, and neither survives a change of measurement. Both probes use the Anthropic model written evaluations~\citep{perez-etal-2023-discovering}. Figure~\ref{fig:probes}(a) shows the sycophancy probe~\citep{sharma2024towards}, scored on the sycophancy splits (NLP survey, PhilPapers, and political typology questions): the branch ranking flips depending on whether agreement is scored by raw log-probability, chat template log-probability, or chat generation, so no stable branch effect survives. Figure~\ref{fig:probes}(b) shows the persona and value probes, drawn from the persona behavior evaluations of the same suite, as a heatmap of branch differences with no consistent sign or direction across rows. Neither supports a broad claim.

Read together, both panels behave like a null: an effect that looks present under one scoring rule reverses under another, and the persona heatmap shows no coherent direction across value dimensions. This is unlike the refusal result, where the $C_{\mathrm{web}}$--$C_{\mathrm{safety}}$ gap is stable across two independent detectors (Appendix~\ref{app:wildguard}). We therefore read the final window effect as specific to refusal plasticity, not a broad values shift, and keep these probes out of the headline claim.

\section{Limitations and Future Work}
\label{app:limitations}

Our study has several limitations. The controlled setting uses 1B parameter OLMo checkpoints as measurement instruments rather than deployment scale assistants, so we cannot test larger models, though the $C_{\mathrm{web}}$ versus $C_{\mathrm{safety}}$ contrast replicates on a different 1B family (Pythia), which argues against an OLMo specific artifact. A further practical constraint is the limited amount of publicly available high quality corpora for constructing final windows, which bounds the range of content interventions we can test. The protection effect we found is bounded: it raises over-refusal, carries a small capability cost, and weakens as the same window becomes a smaller fraction of prior training. As we isolate only a small final window of pretraining, a broader question is whether the full training path, including earlier data exposure and optimizer history, shapes how a checkpoint responds to later training. Our results also suggest treating pretraining and post-training jointly rather than as independent modules, so that upstream data choices are evaluated partly by how they change the stability and plasticity of downstream updates.

\end{document}